%% file: preprint.tex
\documentclass{article}
\usepackage{spconf,amsmath,graphicx}
\usepackage{graphicx}
\usepackage{amsmath}
\usepackage{amssymb}
\usepackage{commath}
\usepackage{enumitem}
\usepackage{tikz}
\usepackage{pgfplots}
\usepgfplotslibrary{groupplots}
\usepackage{multicol}
\usepackage{multirow}
\usepackage[hidelinks]{hyperref}
\usepackage{booktabs,dcolumn}
\usetikzlibrary{math}
\newcolumntype{d}[1]{D{.}{.}{#1}}
\newcommand{\subhead}[1]{\multicolumn{1}{c}{#1}}%

\usepackage{adjustbox}
\usepackage[utf8]{inputenc}
\usepackage{pgfplots}
\pgfplotsset{compat=1.13}
\DeclareUnicodeCharacter{2212}{−}
\usepgfplotslibrary{groupplots,dateplot}
\usetikzlibrary{patterns,shapes.arrows}

\usepackage[capitalise,noabbrev]{cleveref}
\usepackage{siunitx}
\usepackage{hyphenat}

\newcommand{\CE}{F_\text{CE}}
\newcommand{\RE}{F_\text{RE}}
\newcommand{\DE}{F_\text{GE}}

\usepackage{lipsum}

\usepackage{mwe} %

\usepackage[firstpage=true,color=black,placement=top,scale=1,opacity=1,vshift=-1cm]{background}
\SetBgContents{
 \begin{minipage}{1.9\linewidth}
 \small
 \textcopyright 2022 IEEE.  Personal use of this material is permitted.  Permission from IEEE must be obtained for all other uses, in any current or future media, including reprinting/republishing this material for advertising or promotional purposes, creating new collective works, for resale or redistribution to servers or lists, or reuse of any copyrighted component of this work in other works. A definitive version was published in \textbf{2022 IEEE 19th International Symposium on Biomedical Imaging (ISBI)} and is available at \url{https://dx.doi.org/10.1109/ISBI52829.2022.9761575}
 \end{minipage}
 }

\title{A hybrid multi-object segmentation framework with model-based B-splines for microbial single cell analysis}
 \name{Karina Ruzaeva$^{\star \dagger}$ \qquad Katharina Nöh$^{\star}$ \qquad Benjamin Berkels$^{\dagger}$}
 \address{$^{\dagger}$AICES, RWTH Aachen University, Aachen, Germany \\
     $^{\star}$IBG-1: Biotechnology, Forschungszentrum Jülich GmbH, Jülich, Germany}

\begin{document}
\maketitle
\begin{abstract}
In this paper, we propose a hybrid approach for multi-object microbial cell segmentation. The approach combines an ML-based detection with a geometry-aware variational-based segmentation using B-splines that are parametrized based on a geometric model of the cell shape. The detection is done first using YOLOv5. In a second step, each detected cell is segmented individually. Thus, the segmentation only needs to be done on a per-cell basis, which makes it amenable to a variational approach that incorporates prior knowledge on the geometry. Here, the contour of the segmentation is modelled as closed uniform cubic B-spline, whose control points are parametrized using the known cell geometry. Compared to purely ML-based segmentation approaches, which need accurate segmentation maps as training data that are very laborious to produce, our method just needs bounding boxes as training data. Still, the proposed method performs on par with ML-based segmentation approaches usually used in this context. We study the performance of the proposed method on time-lapse microscopy data of \emph{Corynebacterium glutamicum}.
\end{abstract}
\begin{keywords}
B-Splines, microfluidic single cell analysis, cell segmentation
\end{keywords}

\section{Introduction}
\label{sec:intro}
Microfluidic single-cell analysis with time-lapse microscopy is a versatile tool to study cellular processes with spatio-temporal resolution. For instance, the analysis is used to reveal heterogeneity and cell dynamics in microbial populations with regard to growth, gene expression, cell interactions, production, or regulation \cite{Dusny2020,Ortseifen2020}.
The possibility to generate large-scale datasets with automated time-lapse microscopy requires high-throughput single-cell feature extraction. To extract the features, each cell has to be segmented separately.
Thus, robust, efficient and automatic multi-object segmentation for quantitative single-cell characterization is
crucial~\cite{Leygeber2019}.

Most traditional segmentation methods %
are general-pur\-pose approaches and, thus, do not take prior knowledge on object geometry into account. In consequence, such methods may result in object geometries with unreasonable shapes. Going unnoticed, this may lead to biased results, in terms of, e.g., cell area over time. %
In the case of state-of-the-art ML-based instance segmentation methods like Mask R-CNN \cite{He2017}, where object geometries can be learned implicitly, a large amount of training data is required. For microbial systems, benchmark data sets are lacking. One reason is that the creation of training data for segmentation is time-intense and laborious. In particular, in the case of low-resolution and low-signal-to-noise ratio data, it is complicated to annotate images (draw cell outlines) even for domain experts.

Here, we use the biotechnologically relevant rod-shaped soil bacterium \emph{Corynebacterium glutamicum} (\emph{C.~glutamicum}) \cite{Eggeling2005,Baritugo2018} as a model system to propose a solution for the notorious lack of ground truth (GT) data for microbial single cell analysis. In particular, our contribution to single cell image analysis is the development of a hybrid multi-object segmentation approach. Our approach is examined for \emph{C.~glutamicum} segmentation, and is summarized as follows. First, cells are detected based on the real-time detection framework YOLOv5 \cite{Yolov5}. Second, each detected cell is segmented with a variational geometry-aware spline-based segmentation approach. Finally, the transferability of the segmentation approach to other non-rod-shaped bacteria with known geometries is demonstrated. A similar splitting into a detection step and a B-Spline-based segmentation step is introduced in \cite{Mandal_2021}. The main benefits of the proposed approach are easy-to-create training datasets and that the segmentation preserves geometrical features for each segmentation instance by exploiting the available knowledge on cell morphology.

\section{Methods}

We split multi-object segmentation into two steps.
As first step, we use an ML-based detection approach. The main reason for separating the detection step is unlocking the variational segmentation method, which is otherwise infeasible due to the huge number of object instances to be segmented per image or region of interest \cite{Law2011}.
Often, variational multi-object segmentation methods need laborious initialization and/or an interface for user interaction (e.g.\ manually drawing initial object outlines). Thus, automatic initialization for the subsequent segmentation step is very useful \cite{DelgadoGonzalo2013}. Another benefit of the two-stage process is reducing computational time and helping to avoid failed segmentation, caused by enclosing parts of nearby cells. Additionally, the detector is a useful fast cell counter that provides important process metrics for microfluidic experiments \cite{Gruenberger2012}.
Another advantage of the splitting is that the segmentation can be done in parallel for the cells, since each detected bounding box contains only one full cell, which is then processed independently.
Our source code is available on GitHub\footnote{\url{https://github.com/kruzaeva/model_spline_seg}}.
\subsection{Detection}
\label{detect}
As detection framework, we use YOLOv5, being the fastest detection framework available \cite{Yolov5}, whereas other choices are surely possible.
The training and validation data was generated by the synthetic cell renderer CellSium\footnote{\url{https://github.com/modsim/CellSium}}. In addition, a manually labeled real dataset was used. The overall training set size is $141$ images, where $11$ images are manually labeled. %
\subsection{Objective function}
\label{obj}
After the detection, each detected cell is segmented individually using a variational method.
Variational methods formulate the given task, here, finding the cell border ($\mathcal{C}$), as an optimization problem:
$\mathcal{C}= {\arg\min}_{\tilde{\mathcal{C}}}\; F(\tilde{\mathcal{C}})$

In our case, the idea is to define an objective function $F$ that measures how well our conditions on the segmentation are fulfilled and apply optimization procedures to obtain an optimal segmentation of the central cell from the background (and parts of neighboring cells if present (\cref{fig:shape})) in the detection output.
Specifically, the input are the image tiles, cropped according to the detected bounding boxes, where each image tile contains only one full cell, roughly centered in the tile and possibly parts of neighboring cells.
Let $I:\Omega\to\mathbb{R}$ denote the current image tile and $\Omega\subset\mathbb{R}^2$ the corresponding image domain.

Our objective function $F$ is the sum of three terms:
\begin{equation}
\label{eq:objective}
F(\mathcal{C})=\CE(\mathcal{C})+w_{R}\cdot \RE(\mathcal{C})+w_{D}\cdot \DE(\mathcal{C})
\end{equation}
Here, the terms are responsible for edge detection (contour-based term, $\CE$), for enclosing a region with given properties (region-based term, $\RE$), and for selectively enclosing a single object of interest (geodesic distance-based term, $\DE$).
The selection of the weights $w_R,w_D$ is described in \cref{resseg}.

\paragraph*{Contour-based term}

Object boundaries often coincide with edges in images, which are reflected by large image gradients \cite{DelgadoGonzalo2012,DelgadoGonzalo2013}. Hence, our objective $F$ should maximize the average image gradient intensity along the contour.
The corresponding term of the objective function, which encourages the contour to go through regions of high image gradient, is
\begin{equation}
\label{eq:region+energy_cont}
\CE(\mathcal{C})=\frac{1}{\mathrm{len}(\mathcal{C})}\int_\mathcal{C}\frac{1}{(|\nabla I (c)|+\varepsilon)^{k}}\mathrm{d}\mathcal{H}^1(c).
\end{equation}
Here, $|\nabla I|$ is the norm of the image gradient (\cref{fig:shape}c)), $\varepsilon\!=\!0.001$ a regularization parameter to prevent division by zero, $k\!=\!1.5$ an empirically chosen power, $\mathrm{len}(\mathcal{C})$ the length of $\mathcal{C}$ and $\mathcal{H}^1$ the one-dimensional Hausdorff measure.

\paragraph*{Region-based term}
In case of a crowded cell colony (cells touching each other), $\CE$ attracts the contour also to high-intensity gradient regions corresponding to the border of cells in the neighborhood of our target cell. To counter this undesired behavior, we use the additional term $\RE$.
Such a region energy encodes our prior knowledge on the image intensities.
Since the image was recorded in bright field mode, cells have lower intensities than the surrounding background, minimization of the average intensity inside the contour is beneficial.
The corresponding term in the objective function is
\begin{equation}
\label{eq:regionterm}
\RE(\mathcal{C})=\frac{1}{|\mathcal{R}(\mathcal{C})|}\int_{\mathcal{R}(\mathcal{C})}I(x,y)\mathrm{d}(x,y)
\end{equation}
Here, $\mathcal{R}(\mathcal{C})$ denotes the region enclosed by $\mathcal{C}$.

\paragraph*{Geodesic distance term}
Neither $\CE$ nor $\RE$ can distinguish the target cell in our current tile from neighboring cells that may also be visible in the tile. To account for this, we add a second region (geodesic distance) term, which aims to prevent the contour to enclose any point that would have to be connected to a marker $M^T=[M_x,M_y]$ through a high gradient region \cite{Roberts_2018}. Here, the marker needs to be inside the target cell and, to avoid user interaction, is chosen as the center of the bounding box. The geodesic distance map $D$ (\cref{fig:shape}d) is obtained with the geodesic distance transform by solving the Eikonal equation, implemented with raster scan, using the code from \cite{Wang2019}, with \cref{fig:shape}(b) as an input and an empirically chosen image gradient weighting parameter $=0.8$.

\begin{equation}
\label{eq:geodistterm}
\DE(\mathcal{C})=\frac{1}{|R(\mathcal{C})|}\int_{R(\mathcal{C})}D(x,y)\mathrm{d}(x,y)
\end{equation}
Even though $\RE$ and the geodesic distance term $\DE$, seem to be very similar, their combination improved the segmentation performance in our experiments.  

Integrals over $\mathcal{R}(\mathcal{C})$ can be rephrased using the divergence theorem as line integrals over $\mathcal{C}$ \cite{DelgadoGonzalo2013}.
Let $\mathcal{R}\subset\mathbb{R}^2$ be a compact set with piecewise smooth boundary and $\nu=(\nu_1,\nu_2)$ be the outer normal of $\mathcal{R}$.
For $f:\mathbb{R}^2\to\mathbb{R}$ Lebesque integrable, let $f^x(x,y)=\int_0^xf(t,y)\mathrm{d}t$ and $f^y(x,y)=\int_0^yf(x,t)\mathrm{d}t$. Then, for $F:=\frac12(f^x,f^y)$, we get $\operatorname{div}F=f$ and thus
\begin{equation}
\label{eq:AreaToCurveInt}
\int_\mathcal{R} f\mathrm{d}(x,y)=\frac12\int_{\partial\mathcal{R}}(f^x\nu_1+f^y\nu_2)\mathrm{d}\mathcal{H}^1(c).
\end{equation}
Using \eqref{eq:AreaToCurveInt}, we rephrase $\RE$ and $\DE$ as integrals over $\mathcal{C}$.

\subsection{Geometrical model}
\label{model}

\begin{figure}[!htb]
\centering
\resizebox{.25\columnwidth}{!}{%
\input{tikz/model.tex}%
}
\resizebox{.25\columnwidth}{!}{%
\input{tikz/model_example.tex}%
}
\begin{tikzpicture}
  \node[anchor=south west,inner sep=0] (Bild) at (0,0)
    {\includegraphics[width=.227\columnwidth,height=2.455cm]{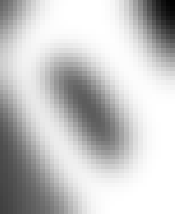}};
  \begin{scope}[x=(Bild.south east),y=(Bild.north west)]
    \node[color=white] at (0.1,0.925) {c)};
  \end{scope}
\end{tikzpicture}
\begin{tikzpicture}
  \node[anchor=south west,inner sep=0] (Bild) at (0,0)
    {\includegraphics[width=.227\columnwidth,height=2.455cm]{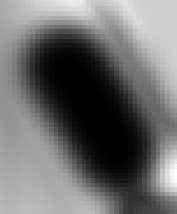}};
  \begin{scope}[x=(Bild.south east),y=(Bild.north west)]
    \node[color=white] at (0.1,0.925) {d)};
  \end{scope}
\end{tikzpicture}

\caption[]{Main components of the analysis: a) Geometrical model of \emph{C.~glutamicum} with $P_i=(P(\theta)_{i1},P(\theta)_{i2})$, b) input of the objective $F$ (cropped image tile $I$ (output of the detection with the proposed model-based segmentation as overlay), c) gradient image $\nabla I$ and d) geodesic distance map $D$).}
\label{fig:shape}
\end{figure}

\emph{C.~glutamicum} cells can be represented with a simple geometrical model, i.e.\ as slightly bent rods. The contour $\mathcal{C}$ is represented as a closed cubic uniform B-Spline curve (for details we refer to \cite{CarldeBoor2001}), as they, in the context of object segmentation, demonstrated solid performance in medical and biological data \cite{DelgadoGonzalo2013,Brigger2000,Mandal2020}.
We propose to exploit the prior knowledge on the geometry by modeling the bent rod shape as closed B-spline curve with $N=6$ control points (cf.\ \cref{fig:shape}), which is parametrized using 8 parameters: length segments ($l_{1}$ and $l_{2}$), width ($w$), 2 curvature parameters ($d$, $e$), center ($c_{x}$, $c_{y}$), and rotation angle $\alpha$. %
Denoting the parameter vector by $\theta=[c_{x},c_{y},l_{1},l_{2},w,d,e,\alpha]$, the resulting coordinates of the six control points $P(\theta)\in\mathbb{R}^{2\times 6}$ are
\begingroup
\belowdisplayskip=0pt
\begin{equation}
\label{eq:model}
P(\theta)=
\begin{bmatrix}
\cos{\alpha}& -\sin{\alpha}\\
\sin{\alpha}&-\cos{\alpha}
\end{bmatrix}
\cdot
\begin{bmatrix}
\tilde{P}_{x}(\theta)-c_{x}\\
\tilde{P}_{y}(\theta)-c_{y}
\end{bmatrix}
+
\begin{bmatrix}
c_{x}\\
c_{y}
\end{bmatrix}
\end{equation}
\endgroup
where
\begin{equation}
\label{eq:modelangle}
\begin{bmatrix}
\tilde{P}_{x}(\theta)\\
\tilde{P}_{y}(\theta)
\end{bmatrix}
\!=\!
\begingroup %
\setlength\arraycolsep{1pt}
\begin{bmatrix}
\frac{w}{2},\!&\!\frac{w}{2}\!-\!d,\!&\!-\frac{w}{2}\!-\!d,\!& -\frac{w}{2},& -\frac{w}{2}\!-\!e, & \frac{w}{2}\!-\!e\\
0, & l_{1},&l_{1},&0,&-l_{2},&-l_{2}\\
\end{bmatrix}
\endgroup
\end{equation}
Note that the idea of parameterizing the geometry is not limited to bent rods, but applies to any kind of prior shape knowledge that can be expressed in terms of a parametrized spline (i.e.\ \cref{result}d)). Such a geometric model-based approach provides target object features (i.e.\ lengths) with no need for post-processing to derive the target parameters from the obtained contour. Moreover, having geometric parameters as variables of an objective function, provides the possibility to apply shape constraints known from the application. In our case, these are biological constraints, such as limits on width $w$ and height $l_{1}\!+\!l_{2}$ of the cells. Considering our rod model, we only expect minor deviations of target object features from the geometric parameters. As a result, the model parameters can be directly considered as target values.%

\subsection{Preprocessing}
To provide better segmentation results, as a preprocessing, we applied a set of simple image processing steps. To avoid undesired cropping of the parts of the cells, a $5$ pixel padding was evenly applied for each bounding box. To unify the range of pixel intensity values of each tile, the intensity image $I$ in \eqref{eq:regionterm}, the gradient image $|\nabla I|$ in \eqref{eq:region+energy_cont} and the geodesic distance map $D$ in \eqref{eq:geodistterm}, were normalized to $[0,1]$. Moreover, to prevent the attraction of the contour by high gradient regions in extracellular space and to prevent erroneous convergence to local minima, we applied Gaussian smoothing ($\sigma=3$) and clipping to $|\nabla I|$ ($[0,0.45]$). This also reduces noise and the influence of undesired internal cell structures.

\subsection{Minimization details}
\paragraph*{Discretization}
To evaluate the integrals over $\mathcal{C}$, an approximation of $\mathcal{C}$ with a discrete curve is needed. To obtain a discrete curve, the contour is evaluated at a finite ($n\!=\!10$) number of equidistant parameter values per spline segment. %
The coordinates of a closed uniform B-spline for a given sequence of parameter values can be expressed as a matrix product of the spline control points coordinates $P_{2 \times N }$ and the discrete spline coordinates matrix $B_{N \times n \cdot N}(t)$, where the matrix $B$ is computed only once per experiment \cite{CarldeBoor2001}.
\paragraph*{Minimization method and constraints}
As discussed in \cref{model}, the organism morphology dependent constraints can be directly applied to the variables of the objective function. According to \cite{Messelink2020}, the segment lengths $l_{1},l_{2}$ are in the range of \SIrange{0.4}{2}{\micro\metre}. The overall length $l\!=\!l_{1}\!+\!l_{2}$ is constrained not to exceed the bounding box diagonal. The deviation from a straight rod in terms of $d,e$ is limited to \SI{0.5}{\micro\metre}. The expected width $w$ of the cell is constrained to \SIrange{0.7}{0.9}{\micro\metre}.
Considering the different ranges of the geometrical parameters ($c_{x}, c_{y}, l_{1}, l_{2}, w, d, e$) and the rotation angle $\alpha$, the objective function $F$ is minimized alternatingly using the ``constrained optimization by linear approximation'' (COBYLA) algorithm \cite{Powell1998}, implemented in \cite{2020SciPy-NMeth}. This means, for a given number of iterations, the function $F$ is minimized with respect to the geometrical parameters for a fixed angle, and analogously, with respect to the angle for fixed geometrical parameters. 
\paragraph*{Initial Guess}
\label{initial_guess}
The initialization may heavily influence the computed solution and the number of iterations, required to reach convergence. To minimize the number of failed segmentations due to undesired local minima, and number of iterations, the initial guess should be chosen carefully. We suggest considering a straight symmetrical ($l_{1}\!=\!l_{2}$) rod with proper orientation as initial guess. Orientation of the rod parameters was chosen considering the bounding box proportions: For a rectangular box (the length of the box exceeds its with by more than 20\%), we use $\alpha\!=\!0^{\circ}$ or $\alpha\!=\!90^{\circ}$, depending on which axis is longer. Otherwise, we use $45^{\circ}$ or $-45^{\circ}$, depending on which angle results in a lower value of $F_{GE}$ value. The initial rod parameters are: $l_{1}\!=\!l_{2}\!=\!0.5*$length of the greater dimension, $w\!=\!17$ pixels, since we have a \SI{0.065}{\micro\metre} pixel size and expect our cells to be around \SI{1}{\micro\metre} wide.

\section{Results}
\label{res}
We evaluated the results of the proposed method using a validation dataset containing, due to the scarcity of real GT data, 30 synthetic (cf.~\cref{detect}) images with $2$ to $196$ cells.
\paragraph*{Detection results}
\label{resdet}
The detection performance was evaluated using the mAP (mean average precision) score at IoU $ =\!0.5$ (intersection over union) and the  mAP averaged over $\mathrm{IoU}\!=\!0.5\!:\!0.95$, cf. \cite{Bell2015}. For the default Yolov5 parameters, $\mathrm{NMS}\!=\!0.45$ (empirically chosen non-maximum suppression parameter, which provides the highest mAP score) and $\mathrm{Confidence}=0.6$, we got:
Average precision $P\!=\!1$;
Average recall $R\!=\!0.98$;
$\mathrm{mAP}_{0.5}\!=\!0.97$;
$\mathrm{mAP}_{0.5:0.95}\!=\!0.76$.
\paragraph*{Segmentation results}
\label{resseg}

We used two metrics to evaluate the segmentation accuracy, based on the Dice score for binary segmentation.
1.\ \textit{Foreground Dice score (FD)}, i.e.\ the dice score of a foreground mask. The cell colony is assumed as union of all single cell masks. Thus, the overlaps of cell masks are treated as belonging to the foreground once.%
2.\ \textit{Average multi-object Dice score (AMD)}, i.e. the average dice score for each single cell in comparison with the corresponding GT mask.
Here, we check the segmentation accuracy, independently of the detection, using GT bounding boxes.
The Dice scores of the proposed constrained geometry-aware method (GA+C) in the \cref{results} is shown against the non-constrained geometry-aware method (GA) and a conventional (non-parametrized) spline fit (nGA), with the same number of control points. For the latter, the objective \eqref{eq:objective} was minimized with respect to control points coordinates. 

The weights $w_R$, $w_D$ were chosen using \cite{Bergstra2013} with $1-\mathrm{AMD}(w_R, w_D)$ as objective function, assuming that $w_R$ and $w_D$ are uniformly distributed in $[0,500]$ with the number of optimization attempts$\!=\!1000$. The objective function was calculated using frame 15 of the validation set sequence, which contains $15$ cells. 
\begin{figure}[!htb]
\centering
\resizebox{\columnwidth}{!}{%
\input{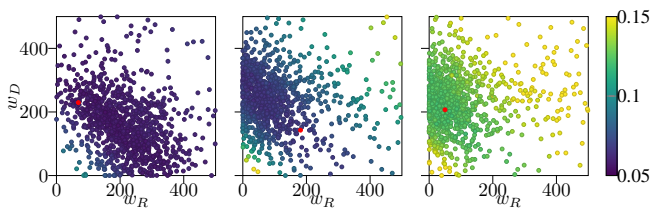}%
}
\caption[]{Weight optimization results. Left to right: geometry-aware segmentation with constraints, without constraints and unconstrained segmentation with control points as variables. The red point shows the best result for each method.}
\label{hyperopt}
\end{figure}
\cref{hyperopt} illustrates, that despite a non-significant gain in terms of accuracy (cf.\ \cref{results}), the constrained geometry-aware segmentation method shows visible improvement in terms of robustness, i.e.\ model insensitivity to the objective function's weights choice. 

The average FD score of the proposed (GA+C) segmentation method, based on the YOLO bounding boxes, for the 30 images, is $0.86$. Unfortunately, a direct comparison with literature results of the state-of-the-art algorithms is not possible since no benchmark data sets with our target microorganism are available. The available numbers indicate that the proposed method outperforms U-net and EDNN based segmentation, trained and tested with comparable amount of real instead of synthetic images \cite{Sachs2018} (0.63 and 0.79, respectively) and is on par with Mask RCNN trained with synthetic data generated by CellSium and tested with real data, in terms of Dice score, applied to a \emph{C.~glutamicum} dataset. Moreover, the proposed method significantly decreases the manual labor for training data creation, since only bounding boxes need to be provided instead of pixel-precise segmentation masks.

\begin{table}
  \centering
  \caption{Segmentation scores. An example GA+C result and the respective ground truth is depicted in \cref{result}a,b}
  \label{results}
  \resizebox{\linewidth}{!}{
  \begin{tabular}{c c c c c c c}
    \toprule 
    \multirow{2}[2]{*}{\shortstack{Image\\ (number of cells)}} & \multicolumn{3}{c}{\textbf{FD}} & \multicolumn{3}{c}{\textbf{AMD}} \\
    \cmidrule(lr){2-4} \cmidrule(lr){5-7} & \subhead{GA+C}&\subhead{GA}&\subhead{nGA}& \subhead{GA+C}& \subhead{GA}& \subhead{nGA}\\ 
    \midrule
         1(2)   & 0.9471& 0.9041 & 0.8982 & 0.9471 & 0.9038 & 0.8978 \\
         5(3)   &  0.9356& 0.9253 & 0.6719 & 0.9445 & 0.9317 & 0.7385 \\
         10(7)  &  0.9420& 0.9391 & 0.8232 & 0.9447 & 0.9405 & 0.8491 \\
         15(15) &  0.9451& 0.9413 & 0.8835 & 0.9460 & 0.9421 & 0.8826 \\
         20(32) &  0.9377& 0.9373 & 0.7036 & 0.9366 & 0.9350 & 0.6375 \\
         25(71) &  0.9408& 0.9386 & 0.6647 & 0.9393 & 0.9355 & 0.6647 \\
         30(196)&  0.9329& 0.9297 & 0.8070 & 0.9297 & 0.9260 & 0.8134 \\
   \bottomrule
 \end{tabular}}
\end{table}
\begin{figure}[!ht]
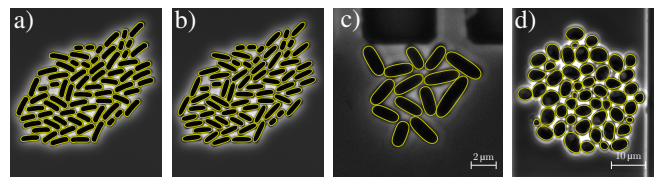

\centering
\resizebox{!}{.26\columnwidth}{%
\input{tikz/gt25.tex}%
}
\resizebox{!}{0.26\columnwidth}{%
\input{tikz/result25.tex}%
}
\resizebox{!}{0.26\columnwidth}{%
\input{tikz/realres.tex}%
}
\resizebox{!}{0.26\columnwidth}{%
\input{tikz/yeast.tex}%
}
\caption[]{a) Ground truth (GT) data and the result of the proposed method applied to b) the GT artificial image, c) real \emph{C.~glutamicum} data based on a rod-shape model (\cref{model}), d) the \emph{S. cerevisiae} based on the ovoid-shape model.}
\label{result}
\vspace{-3ex}
\end{figure}

\section{Conclusions}
We proposed a hybrid approach, combining ML-based detection with variational model-based segmentation. Given the robustness of the approach and easy-to-create training data, we expect it to be an effective framework for other image segmentation tasks in microfluidic single-cell analysis.
The approach is not only limited to rod-shaped cells (\cref{result}c), there are many organisms with known geometries that can be modelled using simple geometrical models. Another example of such an organism is \emph{Saccharomyces cerevisiae}. \emph{S. cerevisiae} cells are round to ovoid, \SIrange{5}{10}{\micro\metre} diameter \cite{Feldmann2010} and can be represented similarly, as a parametrized closed spline with $6$ control points. The result of the geometry-aware segmentation of \emph{S. cerevisiae} is illustrated in \cref{result}d).
\clearpage

\section{Compliance with ethical standards}
\label{sec:ethics}
Ethical approval is not applicable because this work does not contain any studies with animal or human subjects.

\section{Acknowledgments}
\label{sec:acknowledgments}
This work was performed as part of the Helmholtz School for Data Science in Life, Earth and Energy (HDS-LEE) and received funding from the Helmholtz Association.

\bibliographystyle{IEEEbib}
\bibliography{refsetal}

\end{document}

%% file: tikz/model.tex
\begin{tikzpicture}[scale=0.35]
\draw [fill=red] (0,0) circle [radius=0.2] node[above left] {$P_{3}$};
\draw [fill=red] (5,0) circle [radius=0.2] node[above left] {$P_{2}$};
\draw [fill=red] (7,-5) circle [radius=0.2] node[above left] {$P_{1}$};
\draw [fill=red] (5,-12) circle [radius=0.2] node[above left] {$P_{6}$};
\draw [fill=red] (0,-12) circle [radius=0.2] node[above left] {$P_{5}$};
\draw [fill=red] (2,-5) circle [radius=0.2] node[above left] {$P_{4}$};
\draw [fill=gray] (4.5,-5) circle [radius=0.2] node[above] {$c_{x},c_{y}$};
\draw[ thin, ->] (5.5,-5) arc (360:180:1);
\node [below] at (4.5,-6) {$\alpha$};
\draw [thin, <->] (0,0) -- (5,0);
\node [above] at (2.5,0) {$w$};
\draw [thin, <->] (5,0) -- (7,0);
\node [above] at (6,0) {$d$};
\draw [thin, <->] (7,-5) -- (7,0);
\node [right] at (7,-2.5) {$l_{1}$};
\draw [thin, <->] (7,-5) -- (7,-12);
\node [right] at (7,-8.5) {$l_{2}$};
\draw [thin, <->] (5,-12) -- (7,-12);
\node [above] at (6,-12) {$e$};
\node [scale=2.5,color=black] at (-2.2,1) {a)};
\end{tikzpicture}

%% file: tikz/model_example.tex
\begin{tikzpicture}

\definecolor{color0}{rgb}{0.75,0.75,0}

\begin{axis}[
    hide axis,
tick align=outside,
tick pos=left,
axis equal image,
x grid style={white!69.0196078431373!black},
xmin=-0.5, xmax=38.5,
xtick style={color=black},
xtick={-10,0,10,20,30,40},
xticklabels={\ensuremath{-}10,0,10,20,30,40},
y dir=reverse,
y grid style={white!69.0196078431373!black},
ymin=-0.5, ymax=46.5,
ytick style={color=black},
ytick={-10,0,10,20,30,40,50},
yticklabels={\ensuremath{-}10,0,10,20,30,40,50}
]

\addplot graphics [includegraphics cmd=\pgfimage,xmin=-0.5, xmax=38.5, ymin=46.5, ymax=-0.5] {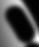};
\addplot [semithick, color0]
table {%
32.702262878418 34.6017761230469
32.2048263549805 35.8827590942383
31.4584274291992 37.0437355041504
30.5041580200195 38.0626792907715
29.3831233978271 38.9175262451172
28.1364212036133 39.586238861084
26.805154800415 40.0467758178711
25.430419921875 40.2770805358887
24.0533199310303 40.2551193237305
22.7149524688721 39.9588394165039
22.7149524688721 39.9588394165039
21.4477443695068 39.3752670288086
20.2494373321533 38.5277328491211
19.1091003417969 37.4486236572266
18.0157985687256 36.1703453063965
16.9585990905762 34.7252883911133
15.9265727996826 33.1458473205566
14.9087820053101 31.4644222259521
13.8942985534668 29.7134094238281
12.8721885681152 27.9252033233643
12.8721885681152 27.9252033233643
11.8393993377686 26.1287899017334
10.8244037628174 24.3395309448242
9.86355876922607 22.5693759918213
8.99321556091309 20.8302745819092
8.24972820281982 19.1341724395752
7.66945171356201 17.4930229187012
7.28874015808105 15.9187717437744
7.14394664764404 14.4233722686768
7.27142572402954 13.0187730789185
7.27142572402954 13.0187730789185
7.69462203979492 11.7186059951782
8.38534450531006 10.5432329177856
9.30249404907227 9.5147008895874
10.4049692153931 8.65505504608154
11.6516704559326 7.98634052276611
13.0014982223511 7.53060340881348
14.4133520126343 7.30988883972168
15.8461322784424 7.34624195098877
17.2587375640869 7.66170930862427
17.2587375640869 7.66170930862427
18.617712020874 8.26899528503418
19.9201564788818 9.14344596862793
21.1708183288574 10.2510652542114
22.3744468688965 11.5578584671021
23.5357837677002 13.0298318862915
24.6595783233643 14.6329870223999
25.7505722045898 16.3333320617676
26.8135166168213 18.0968704223633
27.8531551361084 19.8896083831787
27.8531551361084 19.8896083831787
28.8684196472168 21.6814880371094
29.8349494934082 23.4582214355469
30.7225894927979 25.2094573974609
31.5011653900146 26.9248447418213
32.1405143737793 28.594030380249
32.6104621887207 30.2066688537598
32.8808479309082 31.752405166626
32.9215087890625 33.2208938598633
32.702262878418 34.6017761230469
};
\addplot [semithick, red, mark=*, mark size=3, mark options={solid}, only marks]
table {%
28.1037082672119 19.9543609619141
36.6181640625 35.9241371154785
21.6371994018555 43.9597320556641
13.1227397918701 27.9899559020996
3.1049690246582 11.6316556930542
18.0859375 3.59605979919434
};
\end{axis}
\node [scale=2.5,color=white] at (0.5,5.2) {b)};

\end{tikzpicture}

%% file: tikz/gt25.tex
\begin{tikzpicture}[scale=0.7, every node/.style={inner sep=0,outer sep=0}]

\definecolor{color0}{rgb}{0.75,0.75,0}

\begin{axis}[
    hide axis,
axis equal image,
tick align=outside,
tick pos=left,
x grid style={white!69.0196078431373!black},
xmin=250, xmax=700,
xtick style={color=black},
y grid style={white!69.0196078431373!black},
ymin=400, ymax=900,
ytick style={color=black}
]
\addplot graphics [includegraphics cmd=\pgfimage,xmin=-0.5, xmax=929.5, ymin=1395.5, ymax=-0.5] {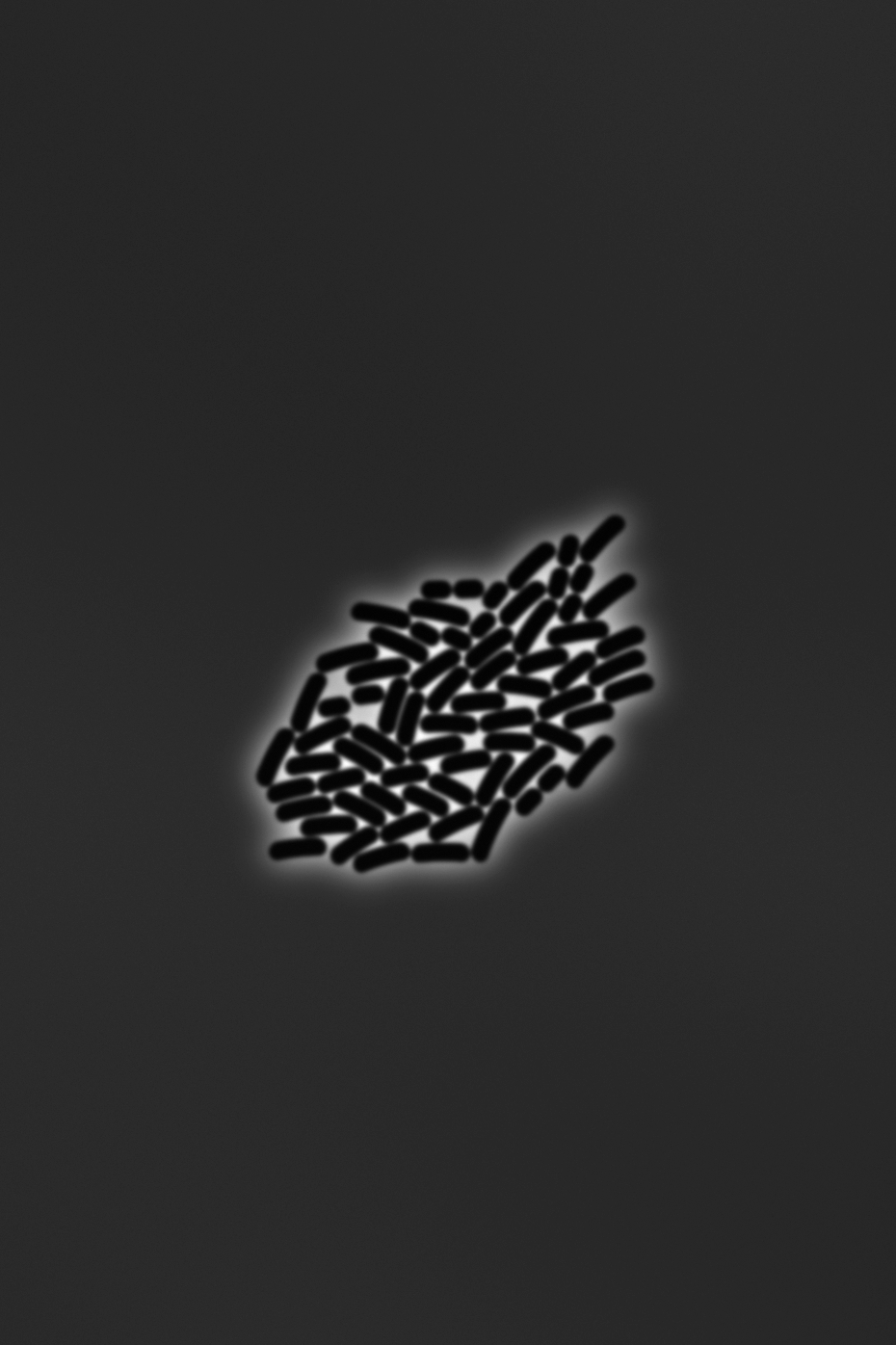};
\node [scale=2.5, color=white] at (290,862) {a)};
\addplot [semithick, color0]
table {%
571 606
570 604
568 603
566 601
564 600
563 598
561 596
559 595
558 593
556 592
554 590
553 588
551 587
549 585
548 583
546 581
545 580
543 578
542 576
540 574
539 572
537 571
537 571
536 569
534 568
531 567
529 567
527 568
525 570
523 572
522 574
522 576
522 579
523 582
524 584
524 584
525 586
527 587
528 589
530 591
531 593
533 595
534 596
536 598
538 600
539 601
541 603
542 605
544 606
546 608
547 610
549 611
551 613
553 614
554 616
556 617
558 619
558 619
560 620
563 621
565 621
568 621
570 620
572 618
574 616
575 614
575 612
574 609
573 607
571 606
};
\addplot [semithick, color0]
table {%
527 552
526 550
525 548
524 546
522 544
521 542
520 540
519 538
518 536
517 534
516 531
515 529
514 527
513 525
512 523
511 521
510 519
510 516
509 514
508 512
507 510
506 507
506 507
505 505
504 503
502 502
500 502
498 502
495 502
493 504
491 506
490 508
489 511
489 513
490 516
490 516
491 518
491 520
492 522
493 525
494 527
495 529
496 531
497 533
498 535
499 538
499 540
501 542
502 544
503 546
504 548
505 550
506 552
507 554
508 556
509 558
511 560
511 560
512 562
514 564
517 565
519 565
522 565
524 564
526 563
527 561
528 559
529 557
528 554
527 552
};
\addplot [semithick, color0]
table {%
575 758
573 756
572 754
570 752
569 751
567 749
566 747
565 745
563 743
562 742
561 740
559 738
558 736
557 734
556 732
554 730
553 728
552 726
551 724
550 722
549 720
549 720
547 718
545 717
543 716
541 716
539 716
537 717
535 719
533 721
533 724
532 726
533 729
533 731
533 731
535 733
536 735
537 737
538 739
539 741
540 743
542 745
543 747
544 749
545 751
547 752
548 754
549 756
551 758
552 760
554 762
555 763
556 765
558 767
559 769
559 769
561 770
563 771
566 772
569 772
571 771
573 770
575 769
576 767
577 764
577 762
576 760
575 758
};
\addplot [semithick, color0]
table {%
429 709
427 710
425 711
423 712
421 713
419 714
417 715
415 716
413 717
411 718
409 719
407 720
404 721
402 722
400 723
398 724
396 724
393 725
391 726
389 727
389 727
387 728
385 729
384 731
383 733
383 736
384 738
385 740
387 742
389 743
392 744
395 744
397 743
397 743
399 743
401 742
404 741
406 740
408 739
410 739
412 738
414 737
417 736
419 735
421 734
423 733
425 732
427 731
429 730
431 729
433 728
435 727
437 725
437 725
439 724
441 722
442 720
442 717
442 715
442 712
440 710
438 709
436 708
434 708
432 708
429 709
};
\addplot [semithick, color0]
table {%
383 709
381 709
379 709
376 708
374 708
372 707
369 707
367 706
365 706
363 705
360 705
358 704
356 704
354 703
351 702
349 702
347 701
345 700
342 699
340 699
340 699
338 698
335 698
333 699
331 700
330 702
329 704
329 706
329 709
330 711
331 713
333 715
335 716
335 716
337 717
340 717
342 718
344 719
346 719
349 720
351 721
353 721
355 722
358 723
360 723
362 724
364 724
367 725
369 725
371 725
374 726
376 726
378 727
378 727
381 727
383 726
386 725
388 723
389 721
390 719
391 717
390 715
389 713
388 711
386 710
383 709
};
\addplot [semithick, color0]
table {%
416 694
414 694
411 694
409 693
407 693
404 693
402 692
400 692
398 692
395 691
393 691
391 691
389 690
386 690
384 689
382 689
380 688
377 687
375 687
373 686
371 686
371 686
368 685
366 685
364 686
362 688
361 689
360 691
360 694
360 696
361 698
363 700
365 702
367 703
367 703
369 704
371 704
374 705
376 705
378 706
380 706
383 707
385 707
387 708
389 708
392 709
394 709
396 709
398 710
401 710
403 710
405 711
408 711
410 711
412 711
412 711
415 711
417 710
420 709
421 708
423 705
424 703
424 701
423 699
422 697
421 696
418 695
416 694
};
\addplot [semithick, color0]
table {%
653 782
651 781
649 779
647 778
645 777
643 775
641 774
640 772
638 771
636 770
634 768
632 767
630 765
628 764
627 762
625 760
623 759
621 757
620 756
620 756
618 754
615 753
613 753
611 753
609 755
607 756
606 758
605 761
605 763
605 766
606 768
608 770
608 770
610 772
611 773
613 775
615 777
617 778
619 780
620 781
622 783
624 784
626 786
628 787
630 789
632 790
634 791
636 793
638 794
640 795
642 797
642 797
644 798
646 798
649 798
652 797
654 796
656 794
657 792
657 790
657 788
657 785
655 783
653 782
};
\addplot [semithick, color0]
table {%
622 733
619 732
617 732
615 732
612 732
610 732
607 731
605 731
603 731
600 730
598 730
595 729
593 729
591 729
588 728
586 728
584 727
581 726
579 726
579 726
576 726
574 726
572 727
570 728
569 730
568 732
568 734
569 737
570 739
571 741
573 742
576 743
576 743
578 744
580 744
583 745
585 745
587 746
590 746
592 747
595 747
597 747
599 748
602 748
604 748
607 749
609 749
611 749
614 749
616 750
619 750
619 750
621 750
624 749
626 748
628 746
629 744
630 742
630 739
629 737
628 735
626 734
624 733
622 733
};
\addplot [semithick, color0]
table {%
474 746
472 747
470 748
467 748
465 749
463 749
460 750
458 750
456 751
453 751
451 752
448 752
446 753
444 753
441 753
439 754
437 754
434 754
432 755
432 755
429 755
427 756
426 758
424 760
424 762
424 764
425 766
426 768
428 770
431 771
433 772
436 772
436 772
438 771
440 771
443 771
445 771
448 770
450 770
452 769
455 769
457 769
459 768
462 768
464 767
467 767
469 766
471 765
474 765
476 764
478 763
478 763
480 762
482 761
484 759
485 757
485 754
485 752
484 750
483 748
481 747
479 746
477 746
474 746
};
\addplot [semithick, color0]
table {%
662 702
660 702
657 701
655 700
653 699
651 698
649 697
647 696
644 696
642 695
640 694
638 693
636 692
634 691
632 689
630 688
628 687
626 686
624 685
624 685
621 684
619 684
617 684
615 685
613 686
612 688
611 691
611 693
611 696
612 698
614 700
616 701
616 701
618 703
620 704
622 705
624 706
626 707
628 708
630 709
632 710
634 711
636 712
638 713
641 714
643 715
645 716
647 716
649 717
651 718
654 719
654 719
656 719
659 719
661 719
663 717
665 716
667 714
667 711
668 709
667 707
666 705
664 703
662 702
};
\addplot [semithick, color0]
table {%
610 666
607 665
605 665
603 664
601 663
599 662
596 662
594 661
592 660
590 659
588 658
586 657
583 656
581 655
579 654
577 653
575 652
573 651
571 650
571 650
569 649
566 649
564 649
562 650
560 652
559 654
558 656
558 659
559 661
560 663
561 665
563 667
563 667
565 668
568 669
570 670
572 671
574 672
576 673
578 674
580 675
582 676
585 677
587 677
589 678
591 679
593 680
595 681
598 681
600 682
602 683
602 683
605 683
607 683
610 682
612 681
614 679
615 677
616 675
616 672
615 670
614 668
612 667
610 666
};
\addplot [semithick, color0]
table {%
336 682
335 680
334 678
333 676
332 674
331 672
330 670
329 668
328 666
327 664
326 661
325 659
324 657
324 655
323 653
322 651
321 648
321 646
320 644
319 642
319 642
318 640
317 638
315 636
313 636
310 636
308 636
306 638
304 640
303 642
302 644
302 647
302 649
302 649
303 652
304 654
305 656
305 658
306 660
307 662
308 665
308 667
309 669
310 671
311 673
312 675
313 677
314 680
315 682
316 684
317 686
318 688
319 690
319 690
320 692
322 693
325 695
327 695
330 695
332 694
334 693
335 691
336 689
337 687
337 685
336 682
};
\addplot [semithick, color0]
table {%
303 625
301 623
300 621
299 619
298 617
297 615
296 613
295 611
294 609
292 607
291 604
290 602
290 600
289 598
288 596
287 594
286 592
285 590
284 588
283 585
283 585
282 583
281 581
279 580
276 580
274 580
272 581
270 582
268 584
267 586
266 589
266 591
267 594
267 594
268 596
269 598
269 600
270 603
271 605
272 607
273 609
274 611
275 613
276 615
277 617
278 619
279 621
280 623
281 625
283 627
284 629
285 631
286 633
286 633
288 635
290 637
292 638
295 638
297 638
299 637
301 635
303 634
304 631
304 629
304 627
303 625
};
\addplot [semithick, color0]
table {%
632 615
631 614
629 612
627 610
626 609
624 607
622 605
621 604
619 602
618 600
616 598
615 597
613 595
612 593
610 591
609 589
608 588
606 586
605 584
603 582
603 582
602 580
600 579
598 578
595 578
593 579
591 580
589 582
588 584
588 587
588 590
588 592
589 594
589 594
591 596
592 598
593 600
595 602
596 604
598 605
599 607
601 609
602 611
604 612
605 614
607 616
608 618
610 619
612 621
613 623
615 624
617 626
618 628
618 628
620 629
623 630
625 630
628 630
630 629
632 628
634 626
635 624
635 621
635 619
634 617
632 615
};
\addplot [semithick, color0]
table {%
405 604
404 605
402 607
400 608
398 609
396 611
394 612
392 613
390 614
388 615
386 617
384 618
382 619
380 620
378 621
376 622
374 623
372 624
370 625
370 625
368 627
366 628
365 630
365 633
365 635
366 637
367 639
370 641
372 642
375 642
377 642
379 641
379 641
382 640
384 639
386 638
388 637
390 636
392 635
394 634
396 632
398 631
400 630
402 629
404 627
406 626
408 625
410 624
412 622
414 621
415 620
415 620
417 618
418 616
419 613
419 611
419 608
418 606
416 604
414 603
412 602
410 602
408 603
405 604
};
\addplot [semithick, color0]
table {%
357 633
355 632
352 631
350 630
348 630
346 629
344 628
342 627
339 626
337 625
335 624
333 623
331 622
329 621
327 619
325 618
323 617
321 616
319 615
319 615
317 614
314 613
312 614
310 615
308 616
307 618
306 620
306 623
306 625
307 628
308 630
310 631
310 631
312 632
314 634
316 635
319 636
321 637
323 638
325 639
327 640
329 641
331 642
333 643
335 644
337 645
340 646
342 647
344 648
346 649
348 649
348 649
351 650
353 650
356 649
358 648
360 646
361 644
362 642
362 640
362 638
361 636
359 634
357 633
};
\addplot [semithick, color0]
table {%
413 743
411 744
408 745
406 745
404 746
401 746
399 747
397 747
394 748
392 748
389 749
387 749
385 750
382 750
380 750
377 751
375 751
373 751
373 751
370 752
368 753
366 754
365 756
365 758
365 760
366 763
367 765
369 766
371 767
374 768
376 768
376 768
379 768
381 768
384 767
386 767
388 767
391 766
393 766
396 765
398 765
400 765
403 764
405 764
407 763
410 762
412 762
414 761
417 761
417 761
419 760
421 758
423 756
424 754
424 752
424 749
423 747
422 745
420 744
418 743
415 743
413 743
};
\addplot [semithick, color0]
table {%
571 815
570 814
568 812
566 811
564 809
563 808
561 806
559 805
558 803
556 802
554 800
553 798
551 797
550 795
548 793
547 792
545 790
544 788
542 787
542 787
540 785
538 784
536 783
534 784
532 784
530 786
528 788
527 790
527 793
527 795
528 798
529 800
529 800
531 802
532 803
534 805
535 807
537 808
538 810
540 812
541 813
543 815
545 817
546 818
548 820
550 821
551 823
553 824
555 826
556 827
558 829
558 829
560 830
563 831
565 831
568 830
570 829
572 828
574 826
574 824
575 821
574 819
573 817
571 815
};
\addplot [semithick, color0]
table {%
382 594
380 595
378 596
376 598
374 599
372 600
370 602
368 603
366 604
364 605
362 606
360 607
358 608
355 610
353 611
351 612
351 612
349 613
347 615
346 617
346 619
346 621
347 623
349 625
351 627
353 628
356 628
358 628
361 627
361 627
363 626
365 625
367 624
369 623
371 622
373 621
376 620
378 619
380 617
382 616
384 615
386 614
388 612
390 611
392 610
392 610
394 608
395 606
396 603
396 601
396 599
395 596
393 594
391 593
389 592
387 592
384 593
382 594
};
\addplot [semithick, color0]
table {%
343 595
341 595
338 595
336 595
333 595
331 595
329 595
326 594
324 594
321 594
319 593
316 593
314 593
312 592
309 592
307 592
307 592
304 591
302 592
300 593
298 594
297 596
297 598
297 600
297 603
298 605
300 606
302 608
305 608
305 608
307 609
309 609
312 610
314 610
317 610
319 611
321 611
324 611
326 611
329 612
331 612
334 612
336 612
338 612
341 612
341 612
343 612
346 611
348 610
350 608
351 606
351 604
351 602
351 600
349 598
348 597
346 596
343 595
};
\addplot [semithick, color0]
table {%
484 634
482 635
479 635
477 635
474 635
472 636
470 636
467 636
465 636
462 636
460 636
457 636
455 636
453 636
450 636
448 636
445 636
445 636
443 636
441 637
439 638
437 639
436 641
436 643
437 646
438 648
439 649
441 651
444 652
446 652
446 652
448 652
451 652
453 653
456 653
458 653
461 653
463 653
465 652
468 652
470 652
473 652
475 652
478 652
480 652
482 651
485 651
485 651
487 651
489 649
491 648
493 646
494 644
494 642
493 639
492 638
491 636
489 635
487 634
484 634
};
\addplot [semithick, color0]
table {%
472 614
470 613
468 613
465 613
463 612
460 612
458 611
456 611
453 610
451 610
449 609
446 609
444 608
442 607
439 607
437 606
435 605
435 605
432 605
430 605
428 606
426 607
424 609
424 611
423 613
424 616
425 618
426 620
428 621
430 622
430 622
433 623
435 624
437 624
439 625
442 626
444 626
446 627
449 627
451 628
454 628
456 629
458 629
461 630
463 630
465 631
468 631
468 631
470 631
473 630
475 629
477 628
479 626
480 624
480 621
479 619
478 617
477 616
475 614
472 614
};
\addplot [semithick, color0]
table {%
530 599
529 597
527 595
526 593
525 591
523 589
522 587
521 585
520 583
519 581
518 579
516 577
515 574
514 572
513 570
512 568
511 566
510 564
510 564
509 562
507 560
505 559
503 559
501 559
499 560
497 561
495 564
494 566
493 568
494 571
494 573
494 573
495 576
496 578
497 580
498 582
499 584
501 586
502 588
503 590
504 592
505 594
506 596
508 598
509 600
510 602
511 604
513 606
514 608
514 608
516 610
518 611
520 612
523 612
525 612
528 611
529 609
531 608
531 605
532 603
531 601
530 599
};
\addplot [semithick, color0]
table {%
495 541
492 540
490 539
488 538
486 538
484 537
481 536
479 535
477 534
475 533
473 532
471 531
469 530
467 529
464 527
462 526
460 525
458 524
458 524
456 523
454 523
451 523
449 524
447 525
446 527
445 530
445 532
446 535
447 537
448 539
450 540
450 540
452 542
454 543
456 544
458 545
460 546
462 547
465 548
467 549
469 550
471 551
473 552
475 553
477 554
480 555
482 556
484 556
486 557
486 557
489 558
491 558
494 557
496 556
498 554
499 552
500 550
500 548
500 546
498 544
497 542
495 541
};
\addplot [semithick, color0]
table {%
472 704
470 703
468 702
466 701
464 699
462 698
460 697
458 696
456 694
454 693
452 692
451 690
449 689
447 688
445 686
443 685
441 683
440 682
440 682
438 681
435 680
433 680
431 680
429 682
427 683
426 686
425 688
425 691
426 693
427 695
429 697
429 697
431 699
432 700
434 701
436 703
438 704
440 706
442 707
444 708
445 709
447 711
449 712
451 713
453 714
455 716
457 717
459 718
461 719
461 719
463 720
466 720
469 720
471 719
473 718
475 716
476 714
476 712
476 709
475 707
474 705
472 704
};
\addplot [semithick, color0]
table {%
421 678
420 676
420 674
419 672
418 670
417 668
416 665
416 663
415 661
414 659
414 657
413 654
412 652
412 650
411 648
411 645
410 643
410 641
410 641
409 638
408 637
406 635
404 634
402 634
399 634
397 635
395 637
394 639
393 642
393 644
393 647
393 647
393 649
394 651
394 653
395 656
395 658
396 660
397 662
397 665
398 667
399 669
399 671
400 673
401 676
402 678
403 680
403 682
404 684
404 684
405 686
407 688
409 690
412 690
414 690
416 690
419 689
420 687
421 685
422 683
422 681
421 678
};
\addplot [semithick, color0]
table {%
385 539
383 540
381 541
379 543
377 544
375 545
373 546
371 547
369 548
366 549
364 550
362 551
360 552
358 553
356 554
354 555
351 556
351 556
349 557
348 559
346 561
346 563
346 565
347 567
348 569
350 571
353 572
355 573
358 573
360 572
360 572
362 571
364 570
367 569
369 568
371 567
373 566
375 565
377 564
379 563
381 562
384 561
386 560
388 559
390 558
392 556
394 555
394 555
396 554
397 552
398 549
398 547
398 544
397 542
396 540
394 539
392 538
389 538
387 538
385 539
};
\addplot [semithick, color0]
table {%
335 551
333 551
330 550
328 550
326 550
323 549
321 549
319 548
316 548
314 548
312 547
309 547
307 546
305 545
302 545
300 544
298 544
298 544
295 543
293 543
291 544
289 546
288 547
287 549
287 552
287 554
288 556
290 558
292 560
294 561
294 561
296 561
299 562
301 563
303 563
305 564
308 564
310 565
312 565
315 566
317 566
319 566
322 567
324 567
326 567
329 568
331 568
331 568
334 568
336 567
338 566
340 564
342 562
343 560
343 558
342 556
341 554
340 552
337 551
335 551
};
\addplot [semithick, color0]
table {%
360 531
358 531
355 531
353 531
350 531
348 531
346 531
343 530
341 530
339 530
336 530
334 530
331 529
329 529
327 529
324 529
322 528
322 528
320 528
317 529
315 530
314 531
312 533
312 535
312 537
313 539
314 541
316 543
318 544
321 545
321 545
323 545
325 546
328 546
330 546
332 546
335 547
337 547
339 547
342 547
344 547
347 547
349 548
351 548
354 548
356 548
359 548
359 548
361 547
363 546
365 545
367 543
368 541
369 539
369 537
368 535
366 533
365 532
362 531
360 531
};
\addplot [semithick, color0]
table {%
328 508
326 507
323 507
321 507
319 507
316 507
314 507
311 507
309 506
307 506
304 506
302 506
300 505
297 505
295 505
293 504
290 504
290 504
288 504
285 504
283 505
282 507
280 508
280 510
280 513
281 515
282 517
284 519
286 520
288 521
288 521
290 521
293 522
295 522
298 522
300 523
302 523
305 523
307 523
309 524
312 524
314 524
316 524
319 524
321 524
324 524
326 524
326 524
329 524
331 523
333 522
335 520
336 518
337 516
336 514
336 512
334 510
333 509
330 508
328 508
};
\addplot [semithick, color0]
table {%
643 843
641 842
640 840
638 839
636 837
635 835
633 834
631 832
630 830
628 828
626 827
625 825
623 823
622 821
620 820
619 818
617 816
617 816
615 814
613 813
611 812
609 813
607 813
605 815
603 817
602 819
602 822
602 824
602 827
604 829
604 829
605 831
607 832
608 834
610 836
611 838
613 840
614 841
616 843
618 845
619 846
621 848
623 850
624 851
626 853
628 855
630 856
630 856
632 858
634 858
637 859
639 858
642 857
644 856
645 854
646 852
646 850
646 847
645 845
643 843
};
\addplot [semithick, color0]
table {%
483 687
481 686
479 684
478 682
476 681
475 679
473 677
471 676
470 674
468 672
467 671
465 669
464 667
462 665
461 664
460 662
458 660
458 660
457 658
455 657
452 657
450 657
448 657
446 659
444 661
443 663
443 665
443 668
443 670
444 673
444 673
446 674
447 676
449 678
450 680
452 681
453 683
455 685
456 687
458 688
459 690
461 692
462 693
464 695
466 696
467 698
469 700
469 700
471 701
473 702
476 702
478 702
481 701
483 700
484 698
485 696
486 693
485 691
484 689
483 687
};
\addplot [semithick, color0]
table {%
438 664
437 662
437 660
436 657
435 655
435 653
434 651
433 649
433 646
432 644
432 642
431 640
430 637
430 635
429 633
429 631
429 628
429 628
428 626
427 624
425 623
423 622
421 621
418 622
416 623
414 624
413 626
412 629
411 631
412 634
412 634
412 636
412 638
413 641
413 643
414 645
415 647
415 650
416 652
416 654
417 656
418 658
418 661
419 663
420 665
420 667
421 669
421 669
422 672
424 673
426 675
428 676
431 676
433 675
435 674
437 673
438 671
439 669
439 666
438 664
};
\addplot [semithick, color0]
table {%
560 671
558 672
555 672
553 673
551 673
548 674
546 674
543 674
541 675
539 675
536 675
534 676
531 676
529 676
527 676
524 676
524 676
522 677
520 678
518 679
517 681
516 683
516 685
517 687
518 689
520 691
522 692
524 693
527 693
527 693
529 693
532 693
534 693
537 693
539 692
541 692
544 692
546 691
549 691
551 691
553 690
556 690
558 689
561 689
563 688
563 688
565 688
567 686
569 684
570 682
571 680
571 678
570 676
569 674
567 672
565 671
563 671
560 671
};
\addplot [semithick, color0]
table {%
514 659
512 659
510 659
507 659
505 659
502 659
500 658
497 658
495 658
493 658
490 658
488 657
485 657
483 657
481 656
478 656
478 656
476 656
473 656
471 657
470 659
469 661
468 663
468 665
469 667
470 669
472 671
474 672
476 673
476 673
479 673
481 674
484 674
486 674
489 674
491 675
493 675
496 675
498 675
501 675
503 675
505 676
508 676
510 676
513 676
513 676
515 675
518 674
520 673
521 671
523 669
523 667
523 665
522 663
521 661
519 660
517 659
514 659
};
\addplot [semithick, color0]
table {%
561 775
559 773
557 772
556 770
554 769
552 767
550 766
549 764
547 763
545 761
543 760
542 758
540 756
538 755
537 753
535 751
534 749
534 749
532 748
530 747
527 746
525 747
523 748
521 749
520 751
519 754
518 756
519 759
519 761
521 763
521 763
522 765
524 766
526 768
527 770
529 771
531 773
532 775
534 776
536 778
538 779
539 781
541 783
543 784
545 786
547 787
548 788
548 788
551 790
553 790
556 790
558 790
560 789
562 787
564 785
564 783
565 781
564 779
563 777
561 775
};
\addplot [semithick, color0]
table {%
527 727
525 725
523 724
521 723
519 722
517 720
515 719
513 717
511 716
509 715
508 713
506 712
504 710
502 709
500 707
499 706
497 704
497 704
495 703
493 702
490 702
488 702
486 703
484 705
483 707
482 710
482 712
483 715
484 717
485 719
485 719
487 720
489 722
491 723
493 725
494 726
496 728
498 729
500 731
502 732
504 733
506 735
507 736
509 738
511 739
513 740
515 741
515 741
518 742
520 743
523 743
525 742
527 741
529 739
530 737
531 735
531 732
530 730
529 728
527 727
};
\addplot [semithick, color0]
table {%
476 502
474 502
472 502
469 502
467 502
465 502
462 502
460 502
458 502
455 502
453 502
451 502
448 502
446 502
444 502
441 502
439 501
437 501
437 501
434 501
432 502
430 503
429 504
428 506
427 508
427 511
428 513
430 515
432 516
434 517
436 518
436 518
439 518
441 518
443 518
446 518
448 518
450 519
453 519
455 519
457 519
460 519
462 519
464 519
467 519
469 519
471 519
474 518
476 518
476 518
478 518
481 517
483 515
484 514
485 511
486 509
485 507
484 505
483 504
481 503
479 502
476 502
};
\addplot [semithick, color0]
table {%
417 502
414 502
412 501
410 501
407 500
405 500
403 499
401 498
399 498
396 497
394 497
392 496
390 495
388 495
385 494
383 493
381 492
379 491
379 491
377 491
374 491
372 491
370 493
368 494
367 496
367 499
367 501
368 503
369 506
371 507
373 508
373 508
376 509
378 510
380 511
382 512
384 512
386 513
389 514
391 514
393 515
395 516
398 516
400 517
402 517
404 518
406 518
409 519
411 519
411 519
414 519
416 519
418 518
420 516
422 514
423 512
424 510
423 508
422 506
421 504
419 503
417 502
};
\addplot [semithick, color0]
table {%
592 614
589 615
587 616
585 617
583 618
581 619
579 620
577 621
575 622
573 623
571 624
568 625
566 626
564 626
562 627
560 628
557 629
557 629
555 630
554 631
552 633
552 635
552 637
552 640
554 642
555 643
558 645
560 645
563 646
565 645
565 645
567 644
569 644
572 643
574 642
576 641
578 640
580 640
582 639
585 638
587 637
589 636
591 635
593 634
595 633
597 632
599 631
599 631
601 629
603 627
604 625
604 623
604 620
604 618
602 616
601 614
598 613
596 613
594 613
592 614
};
\addplot [semithick, color0]
table {%
545 642
543 642
541 642
538 642
536 641
534 641
531 641
529 640
527 640
525 640
522 639
520 639
518 638
515 638
513 637
511 637
509 636
509 636
506 636
504 636
502 637
500 639
499 640
498 642
498 645
498 647
499 649
501 651
503 653
505 653
505 653
508 654
510 654
512 655
514 655
517 656
519 656
521 657
524 657
526 657
528 658
530 658
533 658
535 659
537 659
540 659
542 659
542 659
545 659
547 659
549 657
551 656
552 654
553 651
553 649
553 647
552 645
550 644
548 643
545 642
};
\addplot [semithick, color0]
table {%
439 535
437 534
435 534
433 533
430 532
428 531
426 530
424 530
422 529
420 528
418 527
416 526
414 525
411 524
409 523
407 522
407 522
405 521
403 521
400 521
398 522
397 524
395 526
395 528
395 531
395 533
396 535
398 537
400 539
400 539
402 540
404 541
406 542
408 543
410 543
412 544
415 545
417 546
419 547
421 548
423 549
425 549
427 550
430 551
432 552
432 552
434 552
437 552
439 551
441 550
443 548
444 546
445 544
445 542
444 539
443 538
441 536
439 535
};
\addplot [semithick, color0]
table {%
386 520
384 519
382 517
380 516
378 515
376 514
374 513
372 511
371 510
369 509
367 508
365 506
363 505
361 504
359 502
358 501
358 501
356 500
353 499
351 499
349 500
347 501
345 503
344 505
344 507
344 510
344 512
345 514
347 516
347 516
349 517
351 519
353 520
355 521
356 523
358 524
360 525
362 527
364 528
366 529
368 530
370 531
372 533
374 534
376 535
376 535
378 536
381 536
383 536
385 535
388 533
389 532
390 529
391 527
390 525
390 523
388 521
386 520
};
\addplot [semithick, color0]
table {%
544 616
542 616
539 616
537 616
535 616
532 617
530 617
528 617
525 617
523 617
521 617
518 617
516 617
514 617
511 616
511 616
509 617
506 617
505 619
503 620
502 622
502 624
502 626
503 628
505 630
507 632
509 633
512 633
512 633
514 633
516 633
519 633
521 633
523 633
526 633
528 633
530 633
533 633
535 633
537 633
540 633
542 633
544 632
544 632
547 632
549 631
551 629
552 627
553 625
554 623
553 621
552 619
551 618
549 617
546 616
544 616
};
\addplot [semithick, color0]
table {%
500 598
498 598
496 598
493 598
491 597
489 597
487 596
484 596
482 596
480 595
477 595
475 594
473 594
471 593
468 593
468 593
466 592
463 592
461 593
460 595
458 596
457 599
457 601
458 603
459 605
460 607
462 609
465 610
465 610
467 610
469 611
471 611
474 612
476 612
478 613
481 613
483 613
485 614
487 614
490 615
492 615
494 615
497 615
497 615
499 615
502 615
504 614
506 612
507 610
508 608
508 606
508 603
506 602
505 600
503 599
500 598
};
\addplot [semithick, color0]
table {%
581 705
579 704
576 704
574 703
571 703
569 702
567 702
564 701
562 701
560 700
557 699
555 698
553 698
550 697
548 696
548 696
546 696
543 696
541 696
539 698
538 699
537 701
536 704
537 706
537 708
539 710
541 712
543 713
543 713
545 714
548 715
550 715
552 716
554 717
557 717
559 718
562 719
564 719
566 720
569 720
571 721
573 721
576 722
576 722
578 722
581 722
583 721
585 719
587 717
588 715
588 713
588 711
587 709
585 707
583 706
581 705
};
\addplot [semithick, color0]
table {%
530 703
528 702
526 700
524 699
522 697
520 696
518 695
516 693
514 692
512 690
510 689
509 687
507 685
505 684
503 682
503 682
501 681
499 680
497 680
494 680
492 681
491 683
489 685
489 688
489 690
489 693
490 695
492 697
492 697
493 698
495 700
497 701
499 703
501 705
503 706
505 708
506 709
508 710
510 712
512 713
514 715
516 716
518 717
518 717
520 718
523 719
526 719
528 718
530 717
532 715
533 713
534 711
534 709
533 706
532 704
530 703
};
\addplot [semithick, color0]
table {%
661 726
659 726
657 725
655 724
652 723
650 722
648 721
646 720
644 719
642 718
639 717
637 716
635 715
633 714
631 713
631 713
629 712
626 712
624 712
622 713
620 715
619 717
618 719
618 722
619 724
620 726
621 728
623 730
623 730
625 731
627 732
630 733
632 734
634 735
636 736
638 737
640 738
642 739
645 740
647 740
649 741
651 742
653 743
653 743
656 743
658 743
661 743
663 741
665 740
666 738
667 735
667 733
666 731
665 729
663 728
661 726
};
\addplot [semithick, color0]
table {%
613 702
611 701
609 699
607 698
606 696
604 695
602 694
600 692
598 691
596 689
595 688
593 686
591 684
589 683
588 681
588 681
586 680
584 679
581 679
579 679
577 680
575 682
574 684
573 686
573 689
573 691
574 694
576 696
576 696
578 697
579 699
581 700
583 702
585 703
586 705
588 706
590 708
592 709
594 711
596 712
598 713
599 715
601 716
601 716
604 717
606 718
609 718
611 717
613 716
615 714
616 712
617 710
617 708
616 705
615 704
613 702
};
\addplot [semithick, color0]
table {%
370 580
368 580
365 579
363 579
361 578
359 578
356 577
354 577
352 576
350 576
347 575
345 574
343 574
341 573
341 573
338 573
336 573
334 574
332 575
330 576
330 579
329 581
330 583
330 585
332 587
334 589
336 590
336 590
338 591
340 591
343 592
345 593
347 593
349 594
352 594
354 595
356 595
358 596
361 596
363 597
365 597
365 597
368 597
370 597
372 596
374 594
376 592
377 590
377 588
377 586
376 584
374 582
372 581
370 580
};
\addplot [semithick, color0]
table {%
318 570
316 570
314 569
311 569
309 568
307 568
305 567
302 567
300 566
298 566
296 565
293 565
291 564
289 563
289 563
286 563
284 563
282 564
280 565
279 567
278 569
277 571
278 573
279 576
280 578
282 579
284 580
284 580
287 581
289 581
291 582
293 583
295 583
298 584
300 584
302 585
304 585
307 586
309 586
311 587
314 587
314 587
316 587
319 587
321 585
323 584
324 582
325 580
325 578
325 575
324 574
322 572
320 571
318 570
};
\addplot [semithick, color0]
table {%
476 561
474 562
472 564
470 565
468 566
465 567
463 568
461 569
459 570
457 572
454 573
452 574
450 574
450 574
448 576
446 577
445 579
445 581
445 584
445 586
447 588
449 589
451 591
454 591
456 591
458 590
458 590
461 589
463 588
465 587
467 586
470 585
472 584
474 583
476 582
478 581
481 580
483 578
485 577
485 577
487 576
488 574
489 571
490 569
489 567
488 564
487 562
485 561
483 560
481 560
478 560
476 561
};
\addplot [semithick, color0]
table {%
435 584
433 584
430 584
428 583
425 583
423 583
421 582
418 582
416 581
413 581
411 580
408 580
406 579
406 579
404 579
401 579
399 580
397 581
396 583
395 585
395 588
396 590
397 592
398 594
400 595
403 596
403 596
405 597
407 597
410 598
412 598
415 599
417 599
420 600
422 600
425 600
427 601
429 601
432 601
432 601
434 601
437 600
439 599
441 598
442 596
443 593
443 591
443 589
441 587
440 586
438 585
435 584
};
\addplot [semithick, color0]
table {%
669 678
667 677
665 676
663 676
660 675
658 675
656 674
654 673
651 673
649 672
647 671
645 670
643 670
640 669
638 668
638 668
636 667
633 667
631 668
629 669
628 671
627 673
626 675
626 677
627 680
628 682
630 683
632 685
632 685
634 685
637 686
639 687
641 688
643 689
645 689
648 690
650 691
652 691
654 692
657 693
659 693
661 694
663 694
663 694
666 695
668 694
671 693
673 692
675 690
676 688
676 686
676 684
675 682
674 680
672 678
669 678
};
\addplot [semithick, color0]
table {%
628 647
626 647
623 647
621 646
619 646
616 645
614 644
612 644
610 643
607 643
605 642
603 641
601 641
598 640
596 639
596 639
594 639
591 639
589 639
587 641
586 642
585 644
585 647
585 649
586 651
587 653
589 655
591 656
591 656
593 657
596 658
598 658
600 659
602 660
605 660
607 661
609 661
611 662
614 662
616 663
618 663
621 664
623 664
623 664
625 665
628 664
630 663
632 662
634 660
635 658
635 655
635 653
634 651
632 649
630 648
628 647
};
\addplot [semithick, color0]
table {%
451 550
449 551
447 552
445 553
443 554
440 555
438 556
436 557
434 558
432 559
430 560
428 561
425 562
423 562
423 562
421 564
419 565
418 567
418 569
418 571
419 574
420 576
422 577
424 578
427 579
429 579
431 579
431 579
434 578
436 577
438 576
440 575
442 574
444 573
446 572
449 571
451 570
453 569
455 568
457 567
459 566
459 566
461 564
462 562
463 560
464 558
464 555
463 553
462 551
460 550
458 549
455 548
453 549
451 550
};
\addplot [semithick, color0]
table {%
405 551
403 552
401 553
400 554
398 556
396 557
394 558
392 559
390 560
388 562
385 563
383 564
381 565
379 566
379 566
377 567
376 569
375 571
374 573
375 575
376 578
377 579
379 581
381 582
384 582
386 582
389 581
389 581
391 580
393 579
395 578
397 577
399 576
401 575
403 574
405 572
407 571
409 570
411 569
413 567
415 566
415 566
417 564
418 562
419 560
419 558
419 555
418 553
416 551
414 550
412 549
410 549
408 550
405 551
};
\addplot [semithick, color0]
table {%
613 796
612 795
611 794
611 793
610 791
610 790
609 789
609 788
608 787
607 786
607 786
606 784
605 782
603 781
600 781
598 781
596 782
594 783
593 785
592 787
591 789
591 791
592 794
592 794
593 795
593 796
594 797
594 798
595 800
596 801
596 802
597 803
597 804
597 804
599 806
601 807
603 808
605 809
607 809
609 808
611 806
613 805
613 803
614 800
613 798
613 796
};
\addplot [semithick, color0]
table {%
601 764
600 763
600 762
599 761
598 760
598 759
597 757
597 756
596 755
595 754
595 754
594 752
592 751
590 750
588 749
586 750
584 751
582 752
581 754
580 756
579 758
580 761
580 763
580 763
581 764
582 765
582 766
583 767
584 769
584 770
585 771
586 772
586 773
586 773
588 775
590 776
592 777
594 777
596 777
598 776
600 775
601 773
602 771
602 768
602 766
601 764
};
\addplot [semithick, color0]
table {%
475 722
474 722
473 723
471 723
470 724
469 725
467 725
466 726
465 726
464 727
464 727
462 728
460 730
459 731
458 734
458 736
459 738
460 740
462 741
464 742
467 743
469 743
471 742
471 742
473 742
474 741
475 741
476 740
478 740
479 739
480 738
481 738
483 737
483 737
485 736
486 734
487 732
488 730
488 727
487 725
486 723
484 722
482 721
480 721
477 721
475 722
};
\addplot [semithick, color0]
table {%
441 727
440 727
439 728
438 729
437 729
435 730
434 731
433 731
432 732
430 733
430 733
428 734
427 735
426 737
425 740
426 742
426 744
428 746
430 747
432 748
434 749
436 749
439 748
439 748
440 747
441 747
442 746
444 745
445 745
446 744
447 743
449 743
450 742
450 742
452 741
453 739
454 737
454 734
454 732
453 730
452 728
450 727
448 726
446 726
444 726
441 727
};
\addplot [semithick, color0]
table {%
389 666
387 666
386 666
384 666
383 666
381 666
380 665
378 665
377 665
375 665
375 665
373 665
371 665
369 666
367 668
366 670
365 672
365 674
366 676
367 678
369 680
371 681
373 681
373 681
375 682
376 682
378 682
379 682
381 682
382 683
384 683
385 683
387 683
387 683
389 683
391 682
393 681
395 679
396 677
397 675
397 673
396 671
395 669
393 668
391 667
389 666
};
\addplot [semithick, color0]
table {%
354 654
352 654
351 654
349 653
348 653
347 653
345 653
344 652
342 652
341 652
341 652
338 652
336 652
334 653
332 654
331 656
330 658
330 660
331 663
332 665
333 666
335 668
338 668
338 668
339 669
341 669
342 669
343 669
345 670
346 670
348 670
349 670
351 670
351 670
353 670
356 670
358 669
359 667
361 665
361 663
361 661
361 659
360 657
358 656
356 654
354 654
};
\addplot [semithick, color0]
table {%
523 776
523 775
522 774
521 773
520 772
519 771
519 770
518 769
517 768
517 767
517 767
515 765
513 764
511 764
509 764
507 764
505 765
503 767
502 769
501 771
501 774
502 776
503 778
503 778
504 779
504 780
505 781
506 782
507 783
507 784
508 785
509 786
510 787
510 787
511 788
513 789
516 790
518 790
520 789
522 788
524 786
525 784
525 782
525 780
525 778
523 776
};
\addplot [semithick, color0]
table {%
510 744
509 743
508 742
507 741
506 740
505 740
504 739
504 738
503 737
502 736
502 736
500 735
498 734
496 733
494 734
492 734
490 736
488 738
487 740
487 742
487 745
488 747
490 749
490 749
490 749
491 750
492 751
493 752
494 753
495 754
496 755
497 755
497 756
497 756
499 758
502 758
504 759
506 758
508 757
510 756
512 754
512 752
513 750
512 748
511 746
510 744
};
\addplot [semithick, color0]
table {%
599 827
598 826
598 824
597 823
597 821
597 820
596 819
596 817
596 816
595 814
595 814
594 812
593 810
591 809
590 808
587 808
585 808
583 809
581 811
580 812
579 815
579 817
579 819
579 819
579 821
580 822
580 824
580 825
581 826
581 828
582 829
582 831
582 832
582 832
583 834
585 836
587 837
589 838
591 838
593 838
595 837
597 836
598 834
599 832
599 829
599 827
};
\addplot [semithick, color0]
table {%
588 794
588 792
588 791
587 789
587 788
586 787
586 785
586 784
585 782
585 781
585 781
584 779
583 777
581 775
579 775
577 774
575 775
573 775
571 777
570 779
569 781
568 783
569 786
569 786
569 787
569 788
570 790
570 791
570 793
571 794
571 796
572 797
572 798
572 798
573 801
574 802
576 804
578 805
581 805
583 805
585 804
587 802
588 800
589 798
589 796
588 794
};
\addplot [semithick, color0]
table {%
492 776
491 776
489 776
488 776
487 775
485 775
484 775
483 775
481 775
480 775
480 775
478 775
476 776
474 777
472 778
471 780
471 782
471 785
472 787
473 788
475 790
477 791
479 791
479 791
480 792
482 792
483 792
484 792
486 792
487 792
488 792
490 792
491 792
491 792
493 792
495 791
497 790
499 788
500 786
500 784
500 782
499 780
498 778
496 777
494 776
492 776
};
\addplot [semithick, color0]
table {%
458 775
457 775
455 775
454 775
453 775
452 775
450 775
449 775
448 774
446 774
446 774
444 775
442 775
440 776
439 778
438 780
437 782
437 784
438 786
440 788
441 789
444 790
446 791
446 791
447 791
449 791
450 791
451 791
452 791
454 791
455 791
456 791
458 791
458 791
460 791
462 790
464 789
465 787
466 785
467 783
467 781
466 779
464 777
462 776
460 775
458 775
};
\addplot [semithick, color0]
table {%
582 585
581 584
580 583
579 583
578 582
577 581
576 580
575 579
574 578
573 577
573 577
572 575
570 574
567 574
565 574
563 575
561 576
560 578
559 580
558 583
559 585
559 587
561 589
561 589
562 590
563 591
563 592
564 593
565 594
566 595
567 596
568 597
569 598
569 598
571 599
573 600
575 600
578 600
580 599
582 597
583 596
584 594
584 591
584 589
583 587
582 585
};
\addplot [semithick, color0]
table {%
558 561
557 560
556 559
555 558
554 557
553 556
553 555
552 554
551 553
550 552
550 552
548 551
546 550
544 550
542 550
540 551
538 552
536 554
535 556
535 558
535 561
536 563
537 565
537 565
538 566
539 567
540 568
541 568
542 569
543 570
544 571
545 572
546 573
546 573
547 574
550 575
552 576
554 575
556 574
558 573
560 571
561 569
561 567
560 565
560 563
558 561
};
\end{axis}

\end{tikzpicture}

%% file: tikz/realres.tex
\begin{tikzpicture}[scale=0.9*0.7, every node/.style={inner sep=0,outer sep=0}]

\definecolor{color0}{rgb}{0.75,0.75,0}

\begin{axis}[
    hide axis,
axis equal image,
tick align=outside,
tick pos=left,
x grid style={white!69.0196078431373!black},
xmin=750, xmax=1050,
xtick style={color=black},
y grid style={white!69.0196078431373!black},
ymin=1300, ymax=1600,
ytick style={color=black}
]
\addplot graphics [includegraphics cmd=\pgfimage,xmin=-0.5, xmax=1354.5, ymin=1723.5, ymax=-0.5] {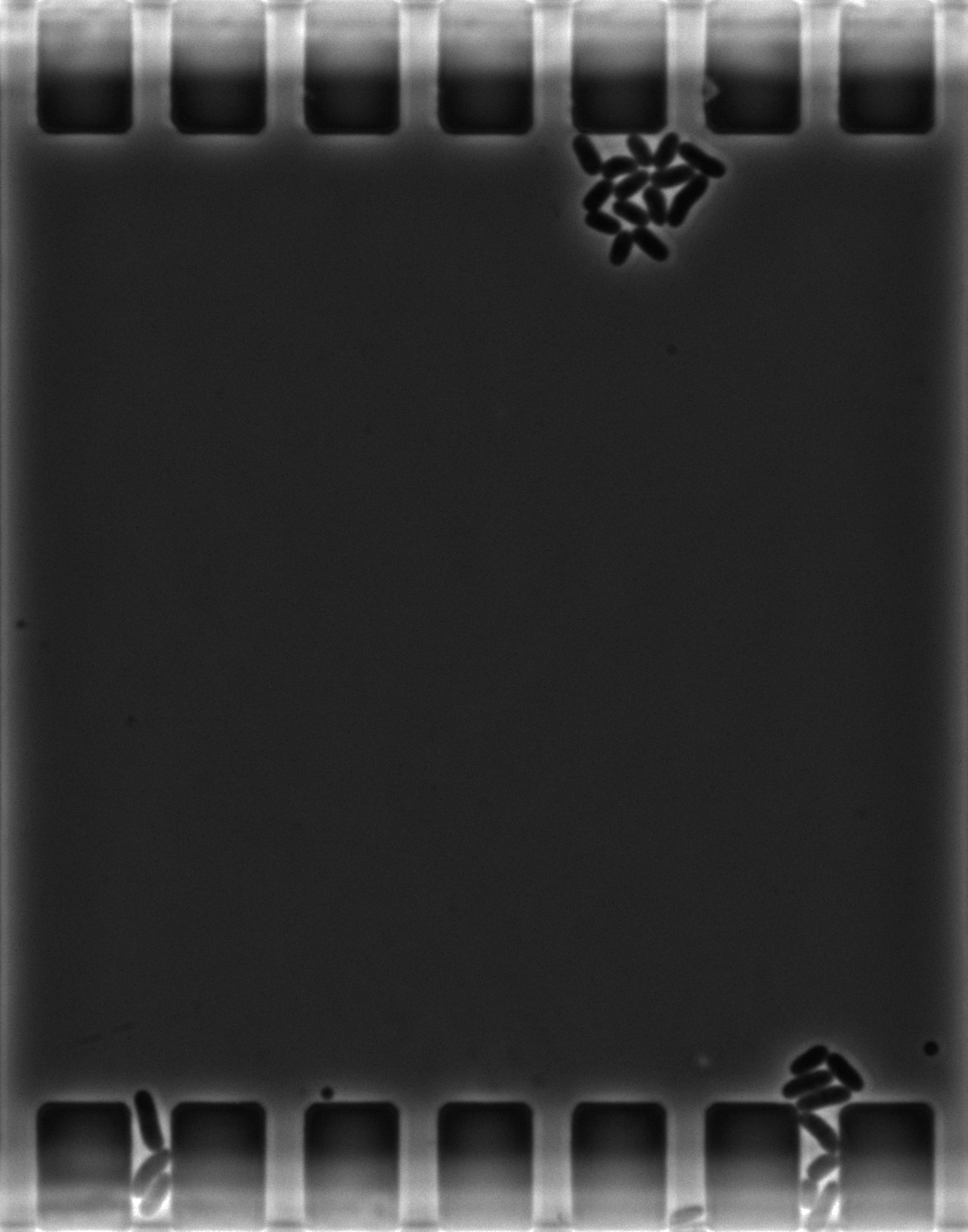};

\draw [thick, white, |-|] (994,1320) -- (1040,1320);
\node [above, white] at (1017,1325) {\SI{2}{\micro\metre}};
\node [scale=2.5, color=white] at (780,1578) {c)};
\addplot [semithick, color0]
table {%
910.080627441406 1498.63781738281
909.568664550781 1496.98010253906
908.757507324219 1495.47985839844
907.694152832031 1494.16552734375
906.425659179688 1493.06530761719
904.9990234375 1492.20751953125
903.461303710938 1491.62048339844
901.859558105469 1491.33227539062
900.240783691406 1491.37145996094
898.652038574219 1491.76611328125
898.652038574219 1491.76611328125
897.130798339844 1492.53210449219
895.676391601562 1493.63488769531
894.27880859375 1495.02746582031
892.927795410156 1496.6630859375
891.613220214844 1498.49450683594
890.324951171875 1500.47497558594
889.052856445312 1502.55737304688
887.786804199219 1504.69482421875
886.5166015625 1506.84033203125
886.5166015625 1506.84033203125
885.24072265625 1508.95385742188
883.992248535156 1511.02270507812
882.812805175781 1513.04113769531
881.743957519531 1515.00354003906
880.827453613281 1516.90417480469
880.104858398438 1518.7373046875
879.617858886719 1520.49719238281
879.408081054688 1522.17810058594
879.517211914062 1523.7744140625
879.517211914062 1523.7744140625
879.972045898438 1525.27648925781
880.740356445312 1526.65991210938
881.775146484375 1527.896484375
883.029418945312 1528.95776367188
884.455993652344 1529.81555175781
886.007995605469 1530.44152832031
887.638305664062 1530.80749511719
889.299865722656 1530.88513183594
890.94580078125 1530.64611816406
890.94580078125 1530.64611816406
892.53759765625 1530.07263183594
894.072082519531 1529.18823242188
895.554565429688 1528.02685546875
896.990478515625 1526.62268066406
898.385131835938 1525.00952148438
899.744018554688 1523.22155761719
901.072387695312 1521.29260253906
902.375793457031 1519.25671386719
903.659484863281 1517.14794921875
903.659484863281 1517.14794921875
904.921813964844 1514.99780273438
906.133056640625 1512.82727050781
907.256225585938 1510.6552734375
908.254455566406 1508.50048828125
909.0908203125 1506.38146972656
909.728576660156 1504.31701660156
910.130676269531 1502.32580566406
910.260314941406 1500.42651367188
910.080627441406 1498.63781738281
};
\addplot [semithick, color0]
table {%
931.790771484375 1414.77722167969
930.816955566406 1413.00415039062
929.537231445312 1411.5205078125
928.011962890625 1410.34301757812
926.3017578125 1409.48889160156
924.466979980469 1408.97497558594
922.568237304688 1408.81811523438
920.666015625 1409.03527832031
918.820678710938 1409.6435546875
917.092834472656 1410.65979003906
917.092834472656 1410.65979003906
915.529357910156 1412.08935546875
914.123474121094 1413.89184570312
912.854736328125 1416.01501464844
911.70263671875 1418.40673828125
910.646911621094 1421.01501464844
909.667114257812 1423.78784179688
908.742858886719 1426.6728515625
907.853698730469 1429.6181640625
906.979248046875 1432.57153320312
906.979248046875 1432.57153320312
906.109497070312 1435.48620605469
905.276062011719 1438.33642578125
904.520935058594 1441.10180664062
903.885986328125 1443.76159667969
903.413330078125 1446.29541015625
903.144836425781 1448.68286132812
903.12255859375 1450.9033203125
903.388488769531 1452.93627929688
903.984497070312 1454.76135253906
903.984497070312 1454.76135253906
904.935607910156 1456.35864257812
906.198303222656 1457.71044921875
907.712219238281 1458.79992675781
909.416748046875 1459.61010742188
911.25146484375 1460.1240234375
913.155883789062 1460.32495117188
915.069519042969 1460.19555664062
916.931884765625 1459.71923828125
918.682495117188 1458.87890625
918.682495117188 1458.87890625
920.274047851562 1457.66662597656
921.711791992188 1456.11083984375
923.014343261719 1454.2490234375
924.2001953125 1452.11865234375
925.287780761719 1449.75708007812
926.295715332031 1447.20166015625
927.242431640625 1444.49011230469
928.146423339844 1441.65966796875
929.026245117188 1438.74768066406
929.026245117188 1438.74768066406
929.890563964844 1435.79150390625
930.709167480469 1432.82641601562
931.441955566406 1429.8876953125
932.048706054688 1427.01049804688
932.489501953125 1424.22998046875
932.724243164062 1421.58117675781
932.712707519531 1419.09936523438
932.414916992188 1416.81970214844
931.790771484375 1414.77722167969
};
\addplot [semithick, color0]
table {%
1011.30963134766 1487.00964355469
1012.00354003906 1484.70739746094
1012.18530273438 1482.4833984375
1011.89208984375 1480.39245605469
1011.16137695312 1478.48901367188
1010.0302734375 1476.828125
1008.53625488281 1475.46447753906
1006.71643066406 1474.45275878906
1004.60815429688 1473.84777832031
1002.24877929688 1473.70422363281
1002.24877929688 1473.70422363281
999.673156738281 1474.05798339844
996.906921386719 1474.86877441406
993.973388671875 1476.07727050781
990.895812988281 1477.62451171875
987.697448730469 1479.451171875
984.401550292969 1481.498046875
981.031433105469 1483.70593261719
977.6103515625 1486.015625
974.161560058594 1488.36804199219
974.161560058594 1488.36804199219
970.71484375 1490.7138671875
967.32568359375 1493.04418945312
964.056213378906 1495.36010742188
960.968444824219 1497.66259765625
958.124328613281 1499.95300292969
955.585998535156 1502.23205566406
953.415466308594 1504.50109863281
951.674682617188 1506.76110839844
950.42578125 1509.01318359375
950.42578125 1509.01318359375
949.714233398438 1511.24914550781
949.519226074219 1513.42333984375
949.803588867188 1515.48132324219
950.529907226562 1517.36804199219
951.660949707031 1519.02893066406
953.159423828125 1520.4091796875
954.988098144531 1521.4541015625
957.109619140625 1522.10876464844
959.486694335938 1522.31860351562
959.486694335938 1522.31860351562
962.084167480469 1522.046875
964.875183105469 1521.3291015625
967.835021972656 1520.21899414062
970.938842773438 1518.77026367188
974.162048339844 1517.03662109375
977.479797363281 1515.07177734375
980.867309570312 1512.92919921875
984.299926757812 1510.66271972656
987.7529296875 1508.32604980469
987.7529296875 1508.32604980469
991.195434570312 1505.96459960938
994.573059082031 1503.59094238281
997.825073242188 1501.20971679688
1000.89099121094 1498.8251953125
1003.71032714844 1496.44189453125
1006.22235107422 1494.06420898438
1008.36663818359 1491.69665527344
1010.08258056641 1489.34362792969
1011.30963134766 1487.00964355469
};
\addplot [semithick, color0]
table {%
846.726623535156 1473.40649414062
848.595153808594 1473.68432617188
850.412170410156 1473.55908203125
852.1318359375 1473.06726074219
853.70849609375 1472.24584960938
855.096313476562 1471.13134765625
856.249633789062 1469.7607421875
857.12255859375 1468.17053222656
857.669494628906 1466.39770507812
857.844543457031 1464.47875976562
857.844543457031 1464.47875976562
857.616577148438 1462.44665527344
857.012573242188 1460.31860351562
856.073974609375 1458.10803222656
854.842346191406 1455.82800292969
853.359130859375 1453.4921875
851.665893554688 1451.11364746094
849.804077148438 1448.70581054688
847.815185546875 1446.28210449219
845.74072265625 1443.85559082031
845.74072265625 1443.85559082031
843.617126464844 1441.44543457031
841.459838867188 1439.09240722656
839.279418945312 1436.84313964844
837.086181640625 1434.74401855469
834.890686035156 1432.84155273438
832.703308105469 1431.18237304688
830.534423828125 1429.81298828125
828.39453125 1428.77966308594
826.294067382812 1428.12927246094
826.294067382812 1428.12927246094
824.249389648438 1427.89404296875
822.300231933594 1428.05151367188
820.492431640625 1428.56457519531
818.871704101562 1429.39672851562
817.48388671875 1430.51123046875
816.374633789062 1431.87109375
815.589782714844 1433.43994140625
815.175048828125 1435.18078613281
815.176147460938 1437.05688476562
815.176147460938 1437.05688476562
815.621887207031 1439.0361328125
816.473022460938 1441.10424804688
817.673461914062 1443.25134277344
819.166931152344 1445.4677734375
820.897277832031 1447.74365234375
822.808349609375 1450.0693359375
824.843933105469 1452.43505859375
826.947814941406 1454.83093261719
829.063842773438 1457.24719238281
829.063842773438 1457.24719238281
831.145874023438 1459.66748046875
833.188110351562 1462.04846191406
835.19482421875 1464.33996582031
837.170227050781 1466.49182128906
839.118591308594 1468.45422363281
841.044189453125 1470.17687988281
842.951171875 1471.60986328125
844.843933105469 1472.703125
846.726623535156 1473.40649414062
};
\addplot [semithick, color0]
table {%
938.533508300781 1532.16625976562
940.218322753906 1532.54772949219
941.886047363281 1532.5400390625
943.492126464844 1532.17565917969
944.992065429688 1531.48693847656
946.341369628906 1530.50610351562
947.49560546875 1529.26550292969
948.410278320312 1527.79748535156
949.040832519531 1526.13452148438
949.342895507812 1524.30883789062
949.342895507812 1524.30883789062
949.285522460938 1522.3505859375
948.891967773438 1520.28173828125
948.199340820312 1518.12182617188
947.244384765625 1515.890625
946.064147949219 1513.60791015625
944.695495605469 1511.29333496094
943.175354003906 1508.96643066406
941.540649414062 1506.64709472656
939.828369140625 1504.35498046875
939.828369140625 1504.35498046875
938.070190429688 1502.11254882812
936.27734375 1499.95336914062
934.455810546875 1497.91394042969
932.611633300781 1496.03063964844
930.750732421875 1494.33972167969
928.879211425781 1492.87780761719
927.003051757812 1491.68115234375
925.128234863281 1490.78625488281
923.2607421875 1490.22937011719
923.2607421875 1490.22937011719
921.4130859375 1490.03552246094
919.623291015625 1490.18395996094
917.935852050781 1490.64208984375
916.395202636719 1491.37780761719
915.0458984375 1492.35864257812
913.932373046875 1493.55236816406
913.099060058594 1494.92651367188
912.590576171875 1496.44873046875
912.451354980469 1498.0869140625
912.451354980469 1498.0869140625
912.7099609375 1499.81323242188
913.331848144531 1501.61889648438
914.266479492188 1503.5
915.463317871094 1505.45239257812
916.871948242188 1507.47192382812
918.44189453125 1509.5546875
920.12255859375 1511.69653320312
921.863464355469 1513.89343261719
923.6142578125 1516.14123535156
923.6142578125 1516.14123535156
925.333984375 1518.42797851562
927.0205078125 1520.70959472656
928.681518554688 1522.93395996094
930.324523925781 1525.04919433594
931.956970214844 1527.00329589844
933.586608886719 1528.74401855469
935.220825195312 1530.21948242188
936.867248535156 1531.37756347656
938.533508300781 1532.16625976562
};
\addplot [semithick, color0]
table {%
869.124267578125 1354.52551269531
867.337158203125 1353.83801269531
865.449035644531 1353.50903320312
863.520324707031 1353.51782226562
861.611267089844 1353.84411621094
859.782104492188 1354.46716308594
858.093200683594 1355.36657714844
856.604858398438 1356.52172851562
855.377380371094 1357.91223144531
854.471069335938 1359.51721191406
854.471069335938 1359.51721191406
853.9287109375 1361.31604003906
853.723449707031 1363.28576660156
853.810913085938 1365.40356445312
854.146667480469 1367.64624023438
854.686340332031 1369.99072265625
855.385559082031 1372.41381835938
856.199951171875 1374.892578125
857.085205078125 1377.40393066406
857.996826171875 1379.92456054688
857.996826171875 1379.92456054688
858.901550292969 1382.42895507812
859.810241699219 1384.88073730469
860.744750976562 1387.24060058594
861.72705078125 1389.46960449219
862.778991699219 1391.52856445312
863.922546386719 1393.37866210938
865.179626464844 1394.98046875
866.572082519531 1396.29516601562
868.121826171875 1397.28344726562
868.121826171875 1397.28344726562
869.837097167969 1397.91650390625
871.671264648438 1398.20458984375
873.564086914062 1398.16845703125
875.455200195312 1397.82849121094
877.284362792969 1397.20544433594
878.9912109375 1396.31958007812
880.515441894531 1395.19177246094
881.796813964844 1393.84228515625
882.775024414062 1392.29187011719
882.775024414062 1392.29187011719
883.406127929688 1390.560546875
883.712158203125 1388.66723632812
883.731506347656 1386.63061523438
883.502502441406 1384.46911621094
883.063598632812 1382.201171875
882.453186035156 1379.84558105469
881.709594726562 1377.42065429688
880.871276855469 1374.94494628906
879.976623535156 1372.43713378906
879.976623535156 1372.43713378906
879.054931640625 1369.91979980469
878.099365234375 1367.43249511719
877.093994140625 1365.01879882812
876.022888183594 1362.72229003906
874.8701171875 1360.58666992188
873.619812011719 1358.65563964844
872.256042480469 1356.97265625
870.762817382812 1355.58142089844
869.124267578125 1354.52551269531
};
\addplot [semithick, color0]
table {%
885.772277832031 1500.88903808594
887.384094238281 1499.94311523438
888.701232910156 1498.74694824219
889.711791992188 1497.35424804688
890.403869628906 1495.81848144531
890.765502929688 1494.193359375
890.784729003906 1492.53247070312
890.449768066406 1490.88916015625
889.74853515625 1489.31726074219
888.669250488281 1487.8701171875
888.669250488281 1487.8701171875
887.209350585938 1486.58911132812
885.404418945312 1485.46557617188
883.299377441406 1484.47839355469
880.939208984375 1483.60668945312
878.368957519531 1482.82934570312
875.633483886719 1482.12536621094
872.777770996094 1481.47387695312
869.846862792969 1480.85363769531
866.8857421875 1480.24389648438
866.8857421875 1480.24389648438
863.936462402344 1479.63293457031
861.030090332031 1479.04736328125
858.194702148438 1478.52355957031
855.458435058594 1478.09729003906
852.849487304688 1477.80493164062
850.39599609375 1477.68237304688
848.126098632812 1477.76599121094
846.067932128906 1478.09167480469
844.249694824219 1478.69555664062
844.249694824219 1478.69555664062
842.696716308594 1479.59899902344
841.423706054688 1480.76318359375
840.442626953125 1482.13452148438
839.765258789062 1483.65954589844
839.403625488281 1485.28479003906
839.369689941406 1486.95629882812
839.675231933594 1488.62084960938
840.332275390625 1490.22485351562
841.352783203125 1491.71447753906
841.352783203125 1491.71447753906
842.739868164062 1493.04821777344
844.462219238281 1494.23156738281
846.479797363281 1495.28198242188
848.752502441406 1496.21704101562
851.240234375 1497.05419921875
853.902954101562 1497.81079101562
856.700561523438 1498.50427246094
859.593078613281 1499.15234375
862.540344238281 1499.77221679688
862.540344238281 1499.77221679688
865.503479003906 1500.373046875
868.448303222656 1500.93078613281
871.341735839844 1501.41259765625
874.150695800781 1501.7861328125
876.842224121094 1502.01879882812
879.383178710938 1502.07800292969
881.740539550781 1501.93115234375
883.881286621094 1501.54565429688
885.772277832031 1500.88903808594
};
\addplot [semithick, color0]
table {%
864.606628417969 1408.87561035156
865.137878417969 1407.24768066406
865.29443359375 1405.58410644531
865.102783203125 1403.93347167969
864.589721679688 1402.34448242188
863.78173828125 1400.86608886719
862.705627441406 1399.546875
861.387878417969 1398.43566894531
859.855224609375 1397.58117675781
858.13427734375 1397.03210449219
858.13427734375 1397.03210449219
856.249816894531 1396.82238769531
854.219543457031 1396.92712402344
852.059387207031 1397.30627441406
849.785095214844 1397.92016601562
847.412536621094 1398.72900390625
844.957641601562 1399.69274902344
842.436218261719 1400.77172851562
839.864196777344 1401.92602539062
837.25732421875 1403.11596679688
837.25732421875 1403.11596679688
834.637023925781 1404.30993652344
832.046630859375 1405.51000976562
829.534851074219 1406.72644042969
827.150573730469 1407.96960449219
824.942626953125 1409.24975585938
822.959716796875 1410.57739257812
821.250793457031 1411.96276855469
819.864501953125 1413.41625976562
818.849792480469 1414.94812011719
818.849792480469 1414.94812011719
818.242797851562 1416.55883789062
818.029479980469 1418.20959472656
818.183227539062 1419.85168457031
818.677368164062 1421.43627929688
819.485290527344 1422.91479492188
820.580383300781 1424.23815917969
821.935974121094 1425.35803222656
823.525451660156 1426.22534179688
825.322143554688 1426.79162597656
825.322143554688 1426.79162597656
827.300231933594 1427.0224609375
829.436706542969 1426.94189453125
831.70947265625 1426.58813476562
834.096313476562 1425.99975585938
836.575073242188 1425.21496582031
839.12353515625 1424.2724609375
841.719665527344 1423.21032714844
844.341186523438 1422.06713867188
846.965881347656 1420.88134765625
846.965881347656 1420.88134765625
849.568298339844 1419.68334960938
852.109313964844 1418.47204589844
854.54638671875 1417.23876953125
856.837036132812 1415.97436523438
858.938720703125 1414.67016601562
860.80908203125 1413.31701660156
862.405456542969 1411.90612792969
863.685485839844 1410.4287109375
864.606628417969 1408.87561035156
};
\addplot [semithick, color0]
table {%
906.376770019531 1419.41247558594
906.981994628906 1417.64501953125
907.181884765625 1415.890625
907.004516601562 1414.1962890625
906.477661132812 1412.609375
905.629455566406 1411.17712402344
904.48779296875 1409.94665527344
903.080627441406 1408.96520996094
901.435974121094 1408.28002929688
899.581726074219 1407.93835449219
899.581726074219 1407.93835449219
897.544921875 1407.97192382812
895.348754882812 1408.35131835938
893.015502929688 1409.03161621094
890.567443847656 1409.96789550781
888.026733398438 1411.115234375
885.415649414062 1412.4287109375
882.756469726562 1413.86328125
880.071472167969 1415.37426757812
877.382873535156 1416.91662597656
877.382873535156 1416.91662597656
874.716491699219 1418.4541015625
872.11279296875 1419.98486328125
869.61572265625 1421.51574707031
867.269287109375 1423.05346679688
865.117492675781 1424.60498046875
863.204345703125 1426.17700195312
861.573791503906 1427.7763671875
860.269897460938 1429.41003417969
859.336608886719 1431.08459472656
859.336608886719 1431.08459472656
858.805969238281 1432.7978515625
858.661926269531 1434.51171875
858.876647949219 1436.17895507812
859.422119140625 1437.75244140625
860.270324707031 1439.18469238281
861.393310546875 1440.4287109375
862.76318359375 1441.43713378906
864.351928710938 1442.16284179688
866.131652832031 1442.55871582031
866.131652832031 1442.55871582031
868.076232910156 1442.59191894531
870.167785644531 1442.28833007812
872.390197753906 1441.68835449219
874.727478027344 1440.83251953125
877.16357421875 1439.76098632812
879.682434082031 1438.51440429688
882.26806640625 1437.13305664062
884.904418945312 1435.65747070312
887.575439453125 1434.12780761719
887.575439453125 1434.12780761719
890.259399414062 1432.57763671875
892.911804199219 1431.01159667969
895.482421875 1429.42736816406
897.921020507812 1427.82275390625
900.177429199219 1426.19543457031
902.201416015625 1424.54296875
903.942749023438 1422.86328125
905.351318359375 1421.15380859375
906.376770019531 1419.41247558594
};
\addplot [semithick, color0]
table {%
902.524841308594 1482.28015136719
904.321411132812 1481.84020996094
905.855712890625 1481.0478515625
907.107238769531 1479.95422363281
908.055358886719 1478.61010742188
908.679504394531 1477.06640625
908.959106445312 1475.37390136719
908.873596191406 1473.58361816406
908.402404785156 1471.74645996094
907.52490234375 1469.91333007812
907.52490234375 1469.91333007812
906.232299804688 1468.12548828125
904.562622070312 1466.38586425781
902.565551757812 1464.68811035156
900.290954589844 1463.02551269531
897.788452148438 1461.3916015625
895.10791015625 1459.77978515625
892.299072265625 1458.18359375
889.41162109375 1456.59643554688
886.495361328125 1455.01184082031
886.495361328125 1455.01184082031
883.595397949219 1453.43176269531
880.738647460938 1451.89343261719
877.947143554688 1450.4423828125
875.243225097656 1449.12426757812
872.649047851562 1447.98498535156
870.186767578125 1447.07019042969
867.878662109375 1446.42541503906
865.746826171875 1446.09655761719
863.813598632812 1446.12915039062
863.813598632812 1446.12915039062
862.100769042969 1446.55297851562
860.629211425781 1447.3330078125
859.419555664062 1448.41857910156
858.492370605469 1449.75866699219
857.868225097656 1451.30236816406
857.567687988281 1452.99890136719
857.611328125 1454.79724121094
858.019714355469 1456.64660644531
858.8134765625 1458.49597167969
858.8134765625 1458.49597167969
860.002563476562 1460.30395507812
861.554809570312 1462.06640625
863.427429199219 1463.78833007812
865.577575683594 1465.47509765625
867.962585449219 1467.13171386719
870.539611816406 1468.763671875
873.265930175781 1470.37585449219
876.098693847656 1471.9736328125
878.995178222656 1473.56213378906
878.995178222656 1473.56213378906
881.914916992188 1475.13830566406
884.826354980469 1476.666015625
887.700378417969 1478.10107421875
890.5078125 1479.39904785156
893.219482421875 1480.51550292969
895.806213378906 1481.40625
898.23876953125 1482.02685546875
900.488037109375 1482.33288574219
902.524841308594 1482.28015136719
};
\addplot [semithick, color0]
table {%
975.430358886719 1477.9423828125
977.985168457031 1478.84973144531
980.401672363281 1479.16613769531
982.624816894531 1478.93188476562
984.599426269531 1478.18701171875
986.270568847656 1476.97143554688
987.5830078125 1475.32543945312
988.481689453125 1473.28894042969
988.911560058594 1470.90209960938
988.817504882812 1468.20495605469
988.817504882812 1468.20495605469
988.164794921875 1465.23583984375
986.999877929688 1462.02563476562
985.3896484375 1458.603515625
983.401000976562 1454.99865722656
981.100708007812 1451.24035644531
978.555725097656 1447.35766601562
975.832763671875 1443.37976074219
972.998779296875 1439.3359375
970.12060546875 1435.25524902344
970.12060546875 1435.25524902344
967.254699707031 1431.17407226562
964.415588378906 1427.15698242188
961.607604980469 1423.27587890625
958.834838867188 1419.6025390625
956.101623535156 1416.20874023438
953.412109375 1413.16625976562
950.770446777344 1410.546875
948.180969238281 1408.42248535156
945.647766113281 1406.86474609375
945.647766113281 1406.86474609375
943.18359375 1405.92700195312
940.835083007812 1405.58752441406
938.657287597656 1405.80651855469
936.705261230469 1406.54382324219
935.0341796875 1407.75927734375
933.699096679688 1409.41296386719
932.755065917969 1411.46472167969
932.257202148438 1413.87451171875
932.260620117188 1416.60217285156
932.260620117188 1416.60217285156
932.801330566406 1419.60913085938
933.839111328125 1422.8623046875
935.314636230469 1426.32983398438
937.168640136719 1429.97998046875
939.341796875 1433.78125
941.774780273438 1437.70166015625
944.408325195312 1441.70971679688
947.183166503906 1445.7734375
950.039916992188 1449.861328125
950.039916992188 1449.861328125
952.92724609375 1453.93530273438
955.825500488281 1457.93249511719
958.722900390625 1461.78344726562
961.607666015625 1465.4189453125
964.468017578125 1468.76989746094
967.292236328125 1471.76696777344
970.068542480469 1474.34094238281
972.78515625 1476.42248535156
975.430358886719 1477.9423828125
};
\addplot [semithick, color0]
table {%
839.960144042969 1488.72314453125
839.491088867188 1486.69287109375
838.6318359375 1484.88232421875
837.438171386719 1483.32312011719
835.966003417969 1482.04711914062
834.271179199219 1481.0859375
832.409606933594 1480.47119140625
830.437194824219 1480.23461914062
828.409729003906 1480.40783691406
826.383117675781 1481.02270507812
826.383117675781 1481.02270507812
824.403381347656 1482.09631347656
822.477355957031 1483.58825683594
820.601989746094 1485.44372558594
818.774169921875 1487.60766601562
816.990905761719 1490.02526855469
815.249145507812 1492.64147949219
813.545837402344 1495.40161132812
811.877868652344 1498.25073242188
810.242309570312 1501.1337890625
810.242309570312 1501.1337890625
808.643676757812 1504.00231933594
807.117431640625 1506.83386230469
805.706604003906 1509.61193847656
804.454284667969 1512.32043457031
803.403442382812 1514.94299316406
802.59716796875 1517.46337890625
802.07861328125 1519.86535644531
801.890747070312 1522.13269042969
802.076599121094 1524.24914550781
802.076599121094 1524.24914550781
802.662841796875 1526.19592285156
803.610046386719 1527.94384765625
804.8623046875 1529.46118164062
806.36376953125 1530.71630859375
808.05859375 1531.67761230469
809.890869140625 1532.31323242188
811.8046875 1532.59155273438
813.744262695312 1532.48095703125
815.653686523438 1531.94958496094
815.653686523438 1531.94958496094
817.488525390625 1530.97924804688
819.250122070312 1529.6044921875
820.951354980469 1527.87316894531
822.60498046875 1525.83337402344
824.223876953125 1523.53295898438
825.82080078125 1521.01989746094
827.408569335938 1518.34216308594
829 1515.54772949219
830.60791015625 1512.68432617188
830.60791015625 1512.68432617188
832.234191894531 1509.79602050781
833.8369140625 1506.91003417969
835.363342285156 1504.04956054688
836.760559082031 1501.23779296875
837.975769042969 1498.498046875
838.956176757812 1495.853515625
839.64892578125 1493.32739257812
840.001159667969 1490.94287109375
839.960144042969 1488.72314453125
};
\addplot [semithick, color0]
table {%
934.690368652344 1369.27624511719
934.843627929688 1367.302734375
934.541442871094 1365.43688964844
933.829711914062 1363.72094726562
932.754455566406 1362.19763183594
931.361511230469 1360.90942382812
929.69677734375 1359.89855957031
927.806274414062 1359.20764160156
925.73583984375 1358.87915039062
923.531433105469 1358.95556640625
923.531433105469 1358.95556640625
921.233459472656 1359.46472167969
918.8603515625 1360.37585449219
916.425109863281 1361.6435546875
913.940612792969 1363.22253417969
911.419860839844 1365.06726074219
908.875732421875 1367.13256835938
906.3212890625 1369.37280273438
903.769348144531 1371.74267578125
901.23291015625 1374.19702148438
901.23291015625 1374.19702148438
898.731018066406 1376.69506835938
896.306945800781 1379.21606445312
894.010131835938 1381.74438476562
891.889892578125 1384.26379394531
889.99560546875 1386.7587890625
888.376708984375 1389.21313476562
887.082580566406 1391.61108398438
886.16259765625 1393.93688964844
885.666137695312 1396.17443847656
885.666137695312 1396.17443847656
885.626647949219 1398.30358886719
886.014221191406 1400.28625488281
886.782775878906 1402.07983398438
887.886535644531 1403.64208984375
889.279479980469 1404.93041992188
890.915710449219 1405.90234375
892.749328613281 1406.51538085938
894.734436035156 1406.72717285156
896.825012207031 1406.4951171875
896.825012207031 1406.4951171875
898.982360839844 1405.79357910156
901.19580078125 1404.6640625
903.4619140625 1403.1650390625
905.777282714844 1401.35485839844
908.138427734375 1399.29174804688
910.541809082031 1397.0341796875
912.984069824219 1394.64050292969
915.461730957031 1392.1689453125
917.971313476562 1389.67797851562
917.971313476562 1389.67797851562
920.500061035156 1387.21655273438
922.998413085938 1384.79711914062
925.407470703125 1382.42236328125
927.668395996094 1380.09521484375
929.722290039062 1377.81872558594
931.510314941406 1375.59558105469
932.973571777344 1373.42883300781
934.05322265625 1371.32141113281
934.690368652344 1369.27624511719
};
\addplot [semithick, color0]
table {%
968.284973144531 1489.32800292969
970.138061523438 1488.40991210938
971.60302734375 1487.21130371094
972.671630859375 1485.78820800781
973.335571289062 1484.19677734375
973.586547851562 1482.49328613281
973.416320800781 1480.73388671875
972.816589355469 1478.974609375
971.779113769531 1477.27172851562
970.295593261719 1475.68139648438
970.295593261719 1475.68139648438
968.368041992188 1474.24719238281
966.03955078125 1472.96252441406
963.363525390625 1471.80859375
960.393310546875 1470.76611328125
957.182250976562 1469.81616210938
953.783752441406 1468.93969726562
950.251220703125 1468.11755371094
946.637939453125 1467.33081054688
942.997436523438 1466.56042480469
942.997436523438 1466.56042480469
939.380126953125 1465.796875
935.825256347656 1465.06860351562
932.369323730469 1464.41369628906
929.048645019531 1463.8701171875
925.899719238281 1463.47583007812
922.958923339844 1463.26892089844
920.262634277344 1463.28747558594
917.847351074219 1463.56945800781
915.749389648438 1464.15295410156
915.749389648438 1464.15295410156
914.000549316406 1465.06018066406
912.613708496094 1466.25061035156
911.597229003906 1467.66833496094
910.959411621094 1469.25695800781
910.708374023438 1470.96044921875
910.8525390625 1472.72265625
911.400146484375 1474.4873046875
912.359436035156 1476.19836425781
913.738708496094 1477.79956054688
913.738708496094 1477.79956054688
915.537414550781 1479.24719238281
917.719665527344 1480.54699707031
920.240783691406 1481.71716308594
923.056091308594 1482.77575683594
926.120910644531 1483.74096679688
929.390563964844 1484.630859375
932.8203125 1485.46362304688
936.365478515625 1486.25744628906
939.9814453125 1487.03039550781
939.9814453125 1487.03039550781
943.623352050781 1487.79138183594
947.246276855469 1488.51257324219
950.804992675781 1489.15686035156
954.254516601562 1489.68701171875
957.549682617188 1490.06604003906
960.645446777344 1490.25671386719
963.49658203125 1490.22204589844
966.058166503906 1489.9248046875
968.284973144531 1489.32800292969
};
\end{axis}

\end{tikzpicture}